\documentclass[acmtog,nonacm]{acmart}
\AtBeginDocument{%
  \providecommand\BibTeX{{%
    \normalfont B\kern-0.5em{\scshape i\kern-0.25em b}\kern-0.8em\TeX}}}

\setcopyright{acmcopyright}
\acmJournal{TOG}
\acmYear{2022}
\acmVolume{41}
\acmNumber{6}
\acmArticle{264}
\acmMonth{12}
\acmDOI{10.1145/3550454.3555493}

\acmPrice{15.00}
\acmISBN{978-1-4503-XXXX-X/18/06}

\acmSubmissionID{373}

\usepackage{pifont}
\usepackage{wrapfig,lipsum,booktabs}
\usepackage{gensymb}
\usepackage{multirow}
\usepackage{array}
\usepackage{float}
\usepackage{subfig}
\usepackage{soul}
\usepackage{wrapfig}
\usepackage{svg}
\usepackage{pdfpages}
\begin{document}

\newcommand{\rmt}[1]{{\vphantom{\hphantom{#1}}}} 

\newcommand{\cc}[1]{{\vphantom{\hphantom{#1}}}} 

\newcommand{\hb}[1]{{\color{cyan}#1}}
\newcommand{\hbc}[1]{{\color{red} \cc{[Hongbo: #1]}}}

\newcommand{\cf}[1]{{\color{orange} #1}}
\newcommand{\cfc}[1]{{\color{orange} \cc{[Chufeng: #1]}}} 

\newcommand{\wanchao}[1]{{\color{blue}#1}}
\newcommand{\wanchaos}[1]{{\color{blue} \cc{[Wanchao: #1]}}}

\newcommand{\yzs}[1]{{\color{purple}#1}}
\newcommand{\lj}[1]{{\color{brown} [Jing: #1]}}
\newcommand{\zhl}[1]{{\color{teal} [Zhouhui: #1]}}
\newcommand{\ac}[1]{{\color{black}#1}}

\newcommand{\datasetName}{DifferSketching}
\title{\datasetName: How Differently Do People Sketch 3D Objects?}

\author{Chufeng Xiao}
\authornote{Authors contributed equally.}
\affiliation{%
  \institution{School of Creative Media, City University of Hong Kong}
  \country{China}
}
\email{chufeng.xiao@my.cityu.edu.hk}

\author{Wanchao Su}
\authornotemark[1]
\affiliation{%
  \institution{School of Creative Media \& Department of Computer Science, City University of Hong Kong}
  \country{China}
}
\email{wanchao.su@cityu.edu.hk}

\author{Jing Liao}
\affiliation{
  \institution{Department of Computer Science, City University of Hong Kong}
  \country{China}
}
\email{jingliao@cityu.edu.hk}

\author{Zhouhui Lian}
\affiliation{%
  \institution{Wangxuan Institute of Computer Technology, Peking University}
  \country{China}
}
\email{lianzhouhui@pku.edu.cn}

\author{Yi-Zhe Song}
\affiliation{%
  \institution{SketchX, CVSSP, University of Surrey}
  \country{UK}
}
\email{y.song@surrey.ac.uk}

\author{Hongbo Fu}
\authornote{Corresponding author.}
\affiliation{%
  \institution{School of Creative Media, City University of Hong Kong}
  \country{China}
}
\email{hongbofu@cityu.edu.hk}


\begin{abstract}
Multiple sketch datasets have been proposed to understand how people draw 3D objects. 
However, such datasets are often of small scale and cover a small set of objects or categories. In addition, these datasets contain freehand sketches mostly from expert users, making it difficult to compare the drawings by expert and novice users, while such comparisons are critical in informing more effective sketch-based interfaces for either user groups. 
These observations motivate us to analyze how differently people with and without adequate drawing skills sketch 3D objects. We invited 70 novice users and 38 expert users to sketch 136 3D objects, which were presented as 362 images rendered from multiple views. This leads to a new dataset of 3,620 freehand multi-view sketches, which are registered with their corresponding 3D objects under certain views. Our dataset is an order of magnitude larger than the existing datasets. We analyze the collected data at three levels, i.e., sketch-level, stroke-level, and pixel-level,
under both spatial and temporal characteristics, and within and across groups of creators. 
We found that the drawings by professionals and novices show significant differences at stroke-level, both intrinsically and extrinsically.
We demonstrate the usefulness of our dataset in two applications: (i) freehand-style sketch synthesis, and (ii) posing it as a potential benchmark for sketch-based 3D reconstruction. 
\ac{Our dataset and code are available at \url{https://chufengxiao.github.io/DifferSketching/}.}
\end{abstract}



\begin{teaserfigure}
  \includegraphics[width=\textwidth]{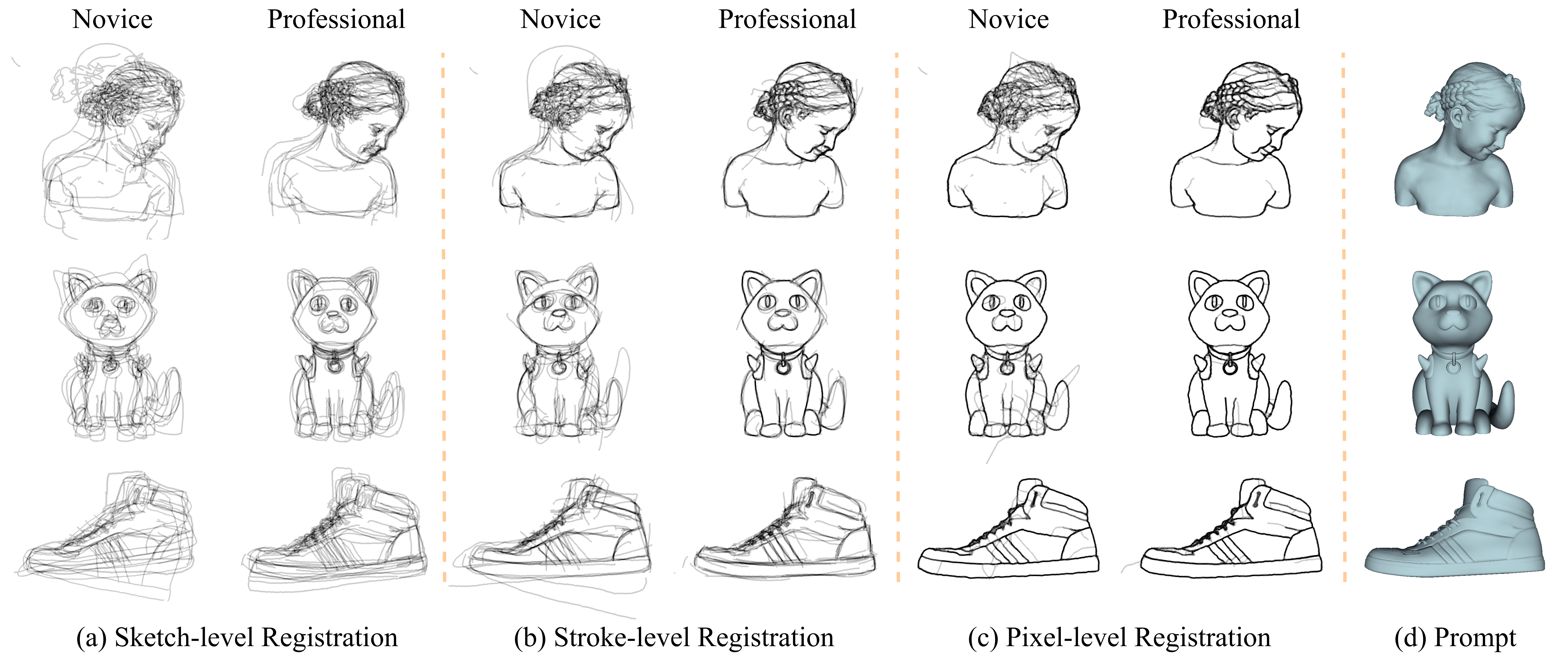}
  \caption{We present \datasetName, a new dataset of freehand sketches to understand how differently professional and novice users sketch 3D objects (d). We perform three-level data analysis through  
  (a) sketch-level registration, (b)  stroke-level registration, and (c) pixel-level registration, to understand the difference in drawings from the two skilled groups.
  }
  \label{fig:teaser}
\end{teaserfigure}

\maketitle

\section{Introduction}\label{sec:introduction}

Using sketches to depict shape and form is a natural and intuitive choice for humans. Consequently, numerous sketch-based interfaces and systems have been proposed for 3D modeling \cite{li2017bendsketch,lun20173d,han2017deepsketch2face} and other content creation tasks.
Many of them carry the claim of being friendly even for \emph{novice users}, i.e., users with limited drawing training/experience. The extent to which such a claim holds is however questionable, since training sketches are predominately algorithm-generated due to a lack of large-scale datasets of paired sketches and 3D models. These pseudo-sketches, albeit being faithful to their corresponding 3D models, do not carry the appropriate level of abstraction and deformation common in novice user sketches. As a result, models trained often carry a strong bias for \emph{professional users}, i.e., users with multiple years of professional drawing training/experience, who are more likely to produce sketches similar to algorithm-generated ones, while performing relatively poorly for novice users, as shown in Section \ref{sec_3d_reconstruction}.

This domain gap between freehand sketches and algorithm-gene-rated line drawings has been known to the community, and was first carefully investigated by Cole et al. \shortcite{cole2008people}. They invited skilled artists to make line drawings given a dozen 3D models and then analyzed these drawings after registering them with the rendered images of the 3D models. More recently, Gryaditskaya et al. \shortcite{gryaditskaya2019opensketch} presented the OpenSketch dataset and by analyzing the collected sketches they aim to understand how professional designers sketch 3D CAD models from new views. 
These datasets lead to insightful findings on how professional artists sketch 3D objects. However, understanding how novice users sketch 3D objects is under explored.
Perhaps more importantly, there is a lack of comprehensive study on how \textit{differently} novice and expert users sketch 3D objects. 
Understanding the differences in how these two diverse groups of people sketch critically informs designs of sketch-based interactive systems that are friendly to both, or specific for either.
Moreover, the aforementioned datasets are of small scale and involve only a dozen 3D objects. Collecting a larger-scale dataset of sketches covering a larger set of objects will not only help derive more general findings but also benefit the training and/or testing of learning-based systems. 

To understand the difference between expert and novice users in drawing 3D objects, we need a new dataset where users with different drawing skills are asked to sketch the same set of 3D objects. A potential solution would be asking users to sketch an imagined 3D object and then asking professional 3D modelers to create the 3D object. The resulting sketch-model pairs might be useful for sketch-based 3D modeling. 
However, sketching upon imagination and modeling upon 2D sketches are both subjective tasks, hindering a comparison between experts' and novices' drawings, which are difficult to register with each other at a stroke level under such a setup. 
Another approach could be sketching a reference object from an unseen viewpoint, as done in \cite{gryaditskaya2019opensketch}. However, this might not work well for novices, since they would not have the same skill to depict a new viewpoint as professionals. 
Aiming for a fair comparison, we follow the data collection approach by Cole et al. \shortcite{cole2008people}, but recruit a larger group of users with different drawing skills and involve a significantly larger set of 3D models from 9 categories. 

Specifically, we invited 70 novice users and 38 expert users to sketch 136 3D objects under 2-3 different views, 
leading to 362 reference images. 
Each reference image was sketched by 5 novice users and 5 expert users. 
We then recruited another group of participants to trace the reference images and utilized the resulting tracings as fiducials to register the freehand drawings of the same reference images. 
By recording the time information for each stroke in the drawing sessions, {we obtained the temporal data along with the freehand sketches, to facilitate further temporal analysis.}
Our study results in a moderately large sketch dataset, involving 3,620 freehand sketches, paired with the corresponding reference images as well as the corresponding 3D models. Our dataset is an order of magnitude larger than the existing datasets for similar purposes (under the stroke-level group in Table \ref{tab_datasets}).

We provide a statistical analysis of the collected freehand sketches in both spatial and temporal perspectives{, and within and across groups of users with similar levels of drawing skills}.
Following a three-level analysis, i.e., sketch-level analysis, stroke-level analysis, and pixel-level analysis (Figure \ref{fig:teaser}), we show the characteristics of the collected sketches and compare their differences according to the drawing skills.
Here we summarize our findings on the similarities and differences of freehand sketches depicted by professionals and novices.
Spatially, we found that the two groups of users 
perceived and intended to draw the same content, while professional users depicted with more precise strokes in terms of intrinsic (shape and form) and extrinsic (scale, rotation, translation) attributes than the novices.
Temporally, both groups of users 
tended to draw the representative regions before depicting the details.
By comparing our collected sketches to widely adopted algorithm-generated drawings, we found that stroke-level registered sketches are potential good proxies for guiding such 
algorithms to generate freehand-like drawings.

Although it seems intuitive to conclude professionals draw better than novices, our analysis focuses on providing quantifiable proofs, benefiting various applications.
We demonstrate how our collected sketches can be used for important applications such as freehand sketch synthesis, and a potential benchmark for sketch-based 3D reconstruction.
With additional annotations, {our collected data can} further serve as new benchmarks for other sketch understanding tasks such as multi-view sketch correspondence {prediction} \cite{yu2020sketchdesc} and semantic segmentation \cite{yang2021sketchgnn}. 
We hope these analysis and insights brought by our new dataset will inspire the design of future sketch-based systems for users with various drawing skills.
We will make the dataset and the code for the synthesis method available to the research community upon the acceptance of this paper.

In summary, this paper makes the following contributions:
\begin{itemize}
 \item A new moderately large dataset of 3,620 sketches of 362 multi-view reference images rendered from 3D models by 70 novice and 38 professional users;
 \item Quantitative comparisons between novice and professional drawings to illustrate the similarities and differences for inspiring the design of sketch-based interfaces and line drawing algorithms; 
 \item Two applications enabled by our dataset: 1) a novel freehand-style sketch synthesis method; 2) a potential benchmark for sketch-based 3D reconstruction.
\end{itemize}
\section{Related Work}
\label{sec:related_work}

In this section, we will mainly focus on discussing the existing sketch collection studies. We will also briefly introduce the extensive efforts towards algorithm-generated line drawings and discuss 
how existing works for learning-based 3D understanding of sketches are trained and evaluated.
Previous works study how people draw differently from the perspectives of perception and cognition.
\ac{For instance, \cite{cohen1997can,calabrese2006influence} investigated the differences in drawings produced by people with various levels of expertise, and such methods evaluated the resulting drawings by letting graders rank them with respect to given aspects. 
These methods provided an empirical analysis of the differences of such drawings, while did not employ absolute measurements. 
To study the differences of drawings from the perspective of elemental components, several methods, e.g., \cite{matthews2008another,schmidt2009expert}, asked their participants to depict specific primitive shapes and analyzed the depiction differences in stroke level.
Tchalenko et al. \shortcite{tchalenko2009segmentation} further investigated the drawing differences by requesting 
participants to copy a long-simple-line artwork and analyzed the differences in stroke level. 
Such analysis-by-collecting methods only provided participants limited targets of drawing, and the accuracy analyses were conducted in stroke level only.
Different to the above mentioned methods, we collected a relative large-scale sketches given 3D models of diverse categories and conducted more comprehensive qualitative and quantitative analysis on them to benefit applications.}

\begin{table*}
\small
    \caption{
    Comparisons of the closely related freehand sketch datasets from various perspectives.
    The ``User'' column indicates the drawing skill levels of the participants involved in the data collection: ``N/A'' means no clear indication or specific requirement of the types of the users; 
    ``P'' and ``N'' represent professional and novice users, respectively. 
    The ``Repetition'' column reports the number of times for each prompt drawn by different users in each dataset.
    In the ``View'' column, ``S'' represents single views mostly used for data collection while ``M'' means the collection of multi-view sketches. 
    ``N/A'' in the ``Category'' column means that there is no clear categorical division in the corresponding dataset.
    Since OpenSketch~\cite{gryaditskaya2019opensketch} collects the design drawings for the CAD models without clear semantic class information, 
    we identify its category as ``N/A''.
    For the data of SpeedTracer \cite{Wang2021Tracing}, we only consider its freehand sketches and exclude its drawings collected via tracing.
    }
    \begin{tabular}{c|c|c|c|c|c|c|c|c|c}
        \hline
        Granularity & Dataset & User & Temporal & Register & Prompt & Repetition & Quantity & Category & View \\
        \hline
        \multirow{3}{4em}{Symbolic}&QuickDraw~\cite{ha2017neural} & N/A & \ding{51} & \ding{55}  & 345 & $\sim$75k & $\sim$25.9M & 345 & S\\
        &SlowSketch~\cite{bhunia2020pixelor} & N/A & \ding{51}  & \ding{55} & 20 & 84-96 & 1,702 & 20 & S\\
        &TU-Berlin~\cite{eitz2012humans} & N/A & \ding{51} & \ding{55} & 250 & $\sim$80 & $\sim$20k & 250 & S \\
        \hline
        \multirow{5}{4em}{Object-level}&Sketchy~\cite{sangkloy2016sketchy} & N/A & \ding{51} & \ding{55}  & 12,500 & $\sim$6 & 75,471 & 125 & S\\
        &FG-SBIR~\cite{yu2016sketch} & N/A & \ding{55} & \ding{55}  & 1,432 & 1 & 1,432 & 2 & S\\
        &{FG-SBSR} ~\cite{qi2021toward} & N/A & \ding{55}  & \ding{55} & 4,680 & 1 & 4,680 & 2 & M\\
         & ShapeNet-Sketch ~\cite{zhang2021sketch2model} & N/A & \ding{55}  & \ding{55} & 1,300 & 1 & 1,300 & 13 & S\\
        &ProSketch-3DChair~\cite{zhong2020towards} & P & \ding{55}  & \ding{55} & 1,500 & 1 & 1,500 & 1 & M\\
        \hline
        \multirow{4}{4em}{Stroke-level}&Princeton~\cite{cole2008people} & P & \ding{55} & \ding{51} & 48 & 1-11 & 208 & N/A & {M} \\
        &OpenSketch~\cite{gryaditskaya2019opensketch} & P & \ding{51} & \ding{51} & 24 & 12-30 & 357 & N/A & M\\
        &SpeedTracer~\cite{Wang2021Tracing} & P & \ding{51} & \ding{51} & 70 & 3-5 & 288 & N/A  & S\\
        &\datasetName ~(Ours) & P+N & \ding{51} & \ding{51} & {362} & 10 & {3,620} & 9 & M\\
        \hline
    \end{tabular}
    
    \label{tab_datasets}
\end{table*}

\begin{figure}[t]
    \centering
    \includegraphics[width=0.9\linewidth]{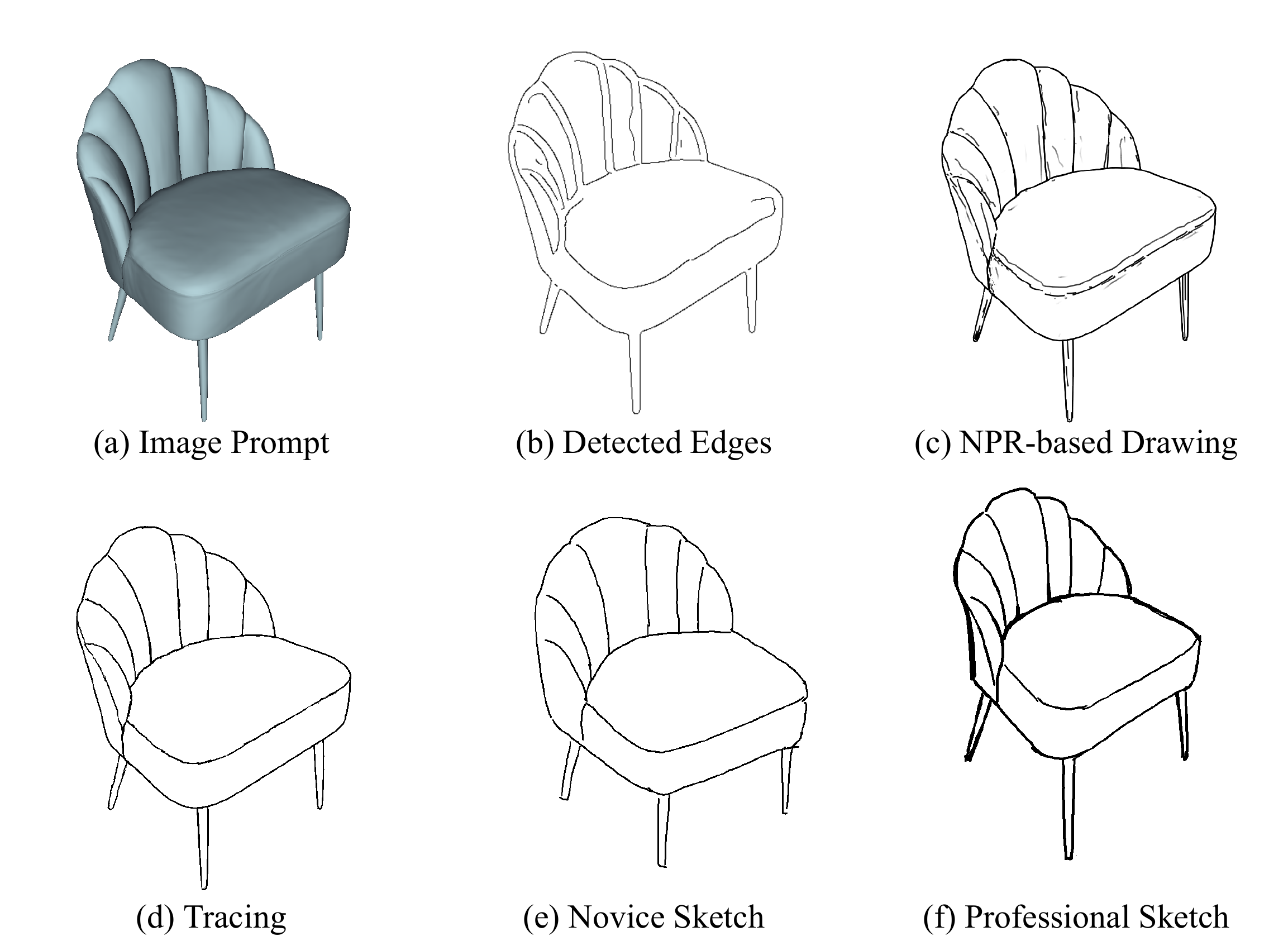}
    \caption{Samples of line drawings generated by different methods. (b) Edges detected by the Canny edge detection algorithm \cite{canny1986computational}. (c) Line drawing produced by Suggestive Contours \cite{decarlo2003suggestive}.
    (d) Manual tracing over (a). (e) From \emph{Novice 11}. (f) From \emph{Professional 2}. 
    }
    \label{fig_diff_mtds}
\end{figure}
\paragraph{Algorithm-generated Line Drawings.} How to generate aesthetically pleasing line drawings from images (e.g., \cite{canny1986computational,bertasius2015deepedge,shen2015deepcontour,liu2017richer}) and 3D models (\cite{decarlo2003suggestive,ohtake2004ridge,judd2007apparent,liu2020neural} has been a long-standing research interest in the community of non-photorealistic rendering. Existing solutions for this task are already able to produce line drawings that are visually pleasing and well aligned with source images or 3D models. They can thus be directly adopted for generating large-scale datasets of pairs of line drawings and images/models, benefiting supervised training of various deep learning-based sketch understanding or sketch-based image/shape generation networks \cite{isola2017image,chenDeepFaceDrawing2020,lun20173d,li2017bendsketch}.
However, as seen from Figure \ref{fig_diff_mtds}, since these algorithms aim to generate line drawings to strictly resemble the geometry features of source images or models, the resulting drawings (in (b) and (c)) are visually different from freehand sketches (in (e) and (f)), especially those by novice users. In fact, freehand sketches often exhibit various levels of abstraction and deformation. This explains why the networks trained on synthesized line drawings often generalize poorly to freehand drawings. Evaluations on synthesized line drawings also could not fully reflect the ability of the trained networks to handle freehand sketches. 
In recent years, a few techniques (e.g., \cite{bhunia2020pixelor,ha2017neural,li2017free}) have also been proposed to synthesize freehand-like sketches. Our collected dataset of paired sketches and tracing can potentially help design more effective sketch synthesis solutions. 

\paragraph{Freehand Sketch Datasets.} The gap between algorithm-generated line drawings and freehand sketches motivated researchers to collect freehand sketches for various tasks such as sketch cleanup \cite{yan2020benchmark} and portrait sketching \cite{berger2013style}. Table~\ref{tab_datasets} summarizes the differences between our \datasetName~and the existing, closely related sketch datasets from various perspectives. These datasets can be roughly categorized into three groups in terms of granularity levels: \emph{symbolic}, \emph{object-level}, and \emph{stroke-level}.  

A sketch dataset under the symbolic group contains sketches describing objects at the category level, {and without reference images. }
Since such sketches are relatively easy to collect, their dataset scale can be very large. Representative datasets under this category include TU-Berlin \cite{eitz2012humans} and QuickDraw \cite{ha2017neural}, which contain thousands or even millions of freehand sketches, collected according to categorical text references via crowd-sourcing platforms or online games. 
The resulting drawings are usually highly abstract and 
only provide symbolic representations of the concerned object categories, instead of individual object instances. They are thus suitable for category-level sketch understanding tasks such as sketch classification \cite{yu2015sketch,li2020sketch} and category-level sketch-based content retrieval \cite{eitz2012sketch}. 

Object-level sketch datasets (e.g., Sketchy \cite{sangkloy2016sketchy}, ShapeNet-Sketch \cite{zhang2021sketch2model}) contain sketches depicting specific object instances as a whole. This category of sketches has been demonstrated {to be useful} for various applications such as fine-grained sketch-based image retrieval (FG-SBIR) and sketch-based 3D modeling. To collect such sketches, objects in photos \cite{sangkloy2016sketchy,yu2016sketch} or rendered images of 3D models \cite{zhong2020towards,zhang2021sketch2model,qi2021toward} are used as prompts, instead of using categorical text references. Such datasets are generally more time-consuming to collect than symbolic sketches, since more strokes are needed to distinguish individual object instances from others. This explains why the scale of object-level sketch datasets is usually not as big as that of symbolic sketch datasets. From Table \ref{tab_datasets}, it can also be seen that the number of times for each prompt drawn by different participants is very small given a large number of prompt numbers. 
Due to the low repetition of drawing and the lack of fine-grained correspondence between individual sketches and their corresponding objects, such datasets are not suitable for in-depth analysis of drawing characteristics.

Sketches in stroke-level datasets have their strokes registered with corresponding prompts and such datasets are thus collected and analyzed to understand how people draw objects. The seminal work by Cole et al.~\shortcite{cole2008people} invited artists to draw a dozen 3D models under 2 views and 2 lighting conditions. Gryaditskaya et al. \shortcite{gryaditskaya2019opensketch} collected OpenSketch through a more challenging task by asking professional designers to draw 3D CAD models in new perspective views given their orthographic views. Wang et al. \shortcite{Wang2021Tracing} followed a data collection process similar to Cole et al.'s but focused on understanding the differences and similarities between tracings and freehand drawings. Our dataset falls into this group. Our data collection and analysis methods are largely inspired by \cite{Wang2021Tracing,cole2008people}. The key differences between our dataset and the above ones are: 1) while the existing datasets contain drawings by professional users only, our dataset consists of carefully registered sketches drawn by both professional and novice users, allowing us to analyze how differently these two groups of users draw 3D objects. 2) Our dataset covers a significantly larger set of 3D models from nine categories and is an order of magnitude larger than the existing ones.

\paragraph{Learning-based 3D Understanding of Sketches.}
Previous works often utilize algorithm-generated line drawings to train and evaluate deep neural networks for various sketch understanding tasks (e.g., \cite{delanoy20183d,lun20173d,su2018interactive,isola2017image}).
One group of such methods (e.g., \cite{li2017bendsketch,lun20173d,su2018interactive}) predict shapes or intermediates from input line drawings with strict correspondence. In other words, such methods treat the inputs as hard constraints and generate shapes strictly corresponding to the inputs. 
Another group of methods adopt indirect reconstruction strategies: they generate shapes by deforming or refining intermediate shape proxies \cite{du2020sanihead,zhang2021sketch2model,han2017deepsketch2face,guillard2021sketch2mesh} to better {approximate} the geometric features 
contained in the input drawings.
As mentioned earlier, the algorithm-generated line drawings strictly resemble the geometry features of source images or models. Training networks on the dataset of such line drawings makes the trained networks perform well on algorithm-generated line drawings. Generally, the larger deviation between testing sketches and algorithm-generated line drawings, the worse performance the trained networks have. Our collected drawings together with the corresponding line drawings can serve as testing data with different levels of difficulties for challenging the trained networks. We show an application of sketch synthesis and believe our dataset as well as our findings through data analysis might provide inspirations for designing new algorithms for synthesizing line drawings with similar natures to the sketches drawn by novice and professional users. Such synthesized sketches would help train the networks to have a better generalization ability for freehand sketches. 
Similar to our intention, Schmidt et al. \shortcite{schmidt2009expert} explored the errors when making foreshortend projections of 3D curves for expert users in a qualitative perceptual study.
We adopt the same strategy of comparing freehand drawings to the accurate references both qualitatively and quantitatively between the novices and the professionals, providing comparisons in more perspectives.
 
\section{Data Collection}
\subsection{Reference Prompting}
As discussed in Section \ref{sec:introduction}, it seems there is no perfect sketch collection approach that exactly reflects how a user draws a desired 3D model in one's mind during the ideation stage and meanwhile enables the stroke-level analysis of sketches by different users for the same objects. To facilitate applications requiring
precise geometric correspondences between sketches and prompts, we adopt an observational drawing setup, following the approach in \cite{cole2008people}. Specifically, 
we provided participants with visual images rendered from 3D models under selected views as prompts and asked participants to depict the key shape details as much as possible\ac{, i.e., participants should make sure their final drawings contain enough shape information to represent given 3D models.}
To provide participants with clear targets to collect stroke-level sketches, we kept the prompt images visible during the drawing process, without trying to use noise maps to blur the perceptions of the targets \cite{sangkloy2016sketchy}.
\ac{Note that the prompt images are not traceable when collecting freehand drawings.}

\begin{table*}[t]
\small
    \caption{Category-level statistics for our \datasetName. We show the basic quantity of the collected dataset.
    }
    \label{tab_details}
    \begin{tabular}{c|c|c|c|c|c|c|c|c|c}
        \hline
        Category & Animal & Animal Head & Chair & Human Face & Industrial Component & Lamp & Primitive & Shoe & Vehicle \\
        \hline
        Models & 10 & 20 & 20 & 6 & 14 & 13 & 10 & 20 & 15 \\
        Views  & 3 & 3 & 3 & 3 & 2-3 & 2-3 & 1-3 & 3 & 3 \\
        References & 30 & 60 & 60 & 18 & 35 & 34 & 20 & 60 & 45 \\
        Freehand Sketches & 300 & 600 & 600 & 180 & 350 & 340 & 200 & 600 & 450   \\
        \hline
    \end{tabular}

\end{table*}

We chose 136 3D models from 9 categories (Table~\ref{tab_details}). We covered the object categories (e.g., \emph{Chair}, \emph{Vehicle}) that were commonly used in existing sketch-based 3D modeling or shape reconstruction works \cite{zhang2021sketch2model,guillard2021sketch2mesh} so that our collected sketches can be easily used for testing the performance of these works. For each category, we included models covering rich geometric shape variations as well as models used in the existing stroke-level sketch datasets \cite{cole2008people,Wang2021Tracing}. Each model was rendered under 2 or 3 representative viewpoints (except for the sphere model with 1 viewpoint in the \emph{Primitive} category) in perspective projection, leading to in total 362 prompt reference images.
For categories with 
\ac{consistent semantic parts} (e.g., Chair, Vehicle), we \ac{first normalized their scales, aligned their semantics to the same axes, and then} chose canonical views (front- and side-view), while for categories with free forms of shapes, i.e., \ac{those without similar structures} (e.g., Industrial Components, Primitives), we chose other representative views.
We generated the prompt images using \emph{trimesh2} library. The models were rendered using a diffuse cyan material under the Phong shader.
For the categories (e.g. \emph{Chair}, \emph{Vehicle}, \emph{Animal Head}) containing samples with axis-aligned models sharing similar structures, we rendered the representative views using the same viewpoint settings. In contrast, for those with large variations (like \emph{Industrial Component}), we manually select the views using a simple interactive interface for camera manipulation. 
More details for the reference prompt images can be found in Table~\ref{tab_details}. 

\subsection{Sketching Constraints}
Since we would like participants to capture the shape information, we requested participants to only depict the shape details and ignore the lighting/shading effects presented in the reference prompts.
We did not set an exact time limit for each prompt while suggested each participant to limit the time for making each drawing within 10 minutes. 
We first gave the overall perception of each target model that participants were going to depict by showing an overall 3D view of the target using 3D viewing software. 
Then we asked participants to draw the rendered version of the 3D model with line drawings to capture the geometric details at their best. 

To ensure the sketching quality and eliminate the variances caused by different drawing tools, we did not collect our dataset via online crowdsouring platforms, as done in \cite{eitz2012humans}.
Instead, we invited the participants to our laboratory and asked them to make drawings 
on the same drawing device we provided.
Our data collection was conducted in individual sessions, each of which lasted around 1 hour.  
To ensure the consistency of multi-view sketches from the same participants, each participant was required to draw all the assigned reference images during the session, though they might take optional short breaks. 
Each participant in one session was compensated with a cash coupon.

\subsection{User Interface and Device} 
\begin{figure}
    \centering
    \includegraphics[width=0.9\linewidth]{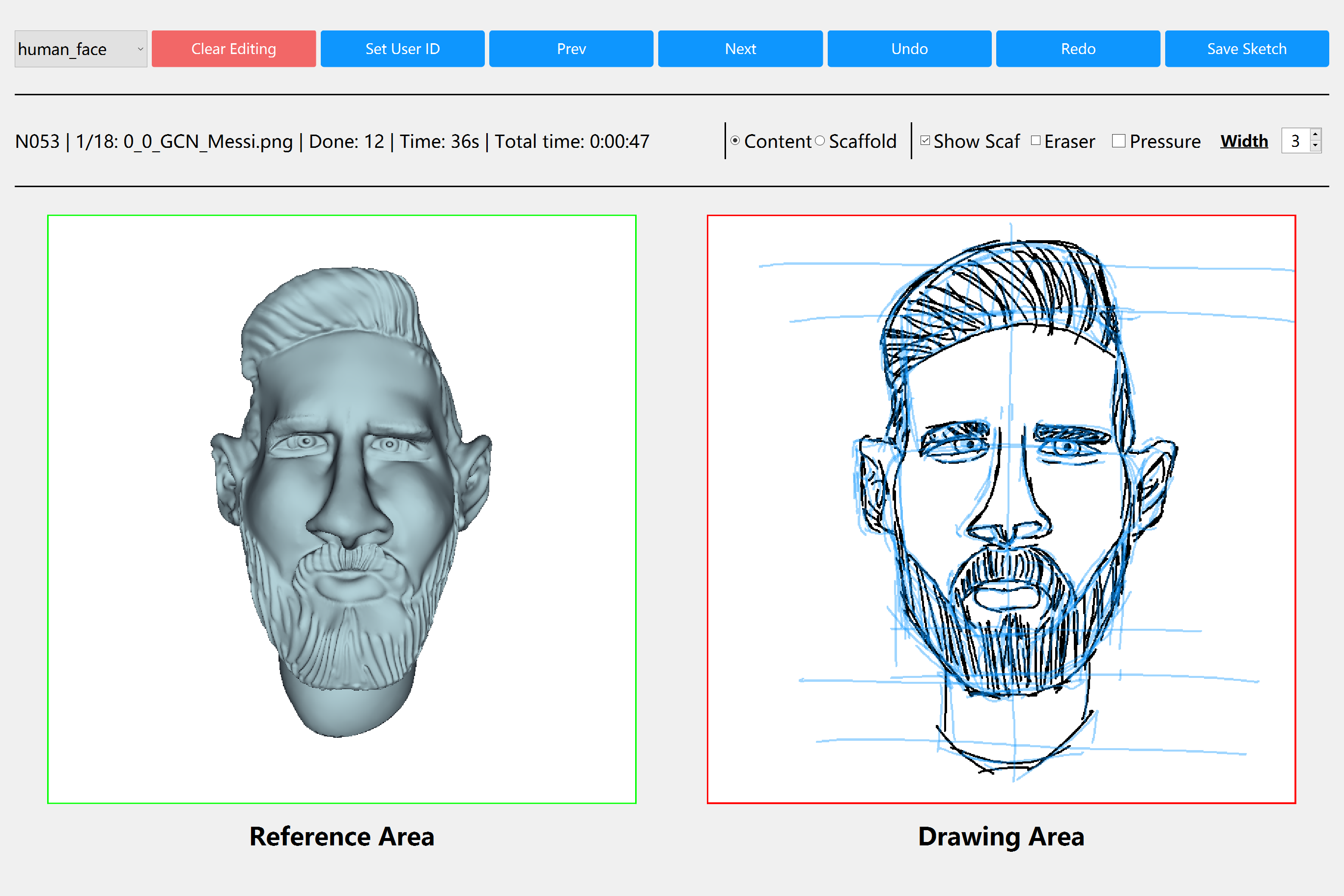}
    \caption{A screenshot of our data collection interface.
    {The left and right panels show a reference image and a user-drawn sketch (from \textit{Novice 53} for this example), respectively.} 
    }
    \label{fig_interface}
\end{figure}
We developed a simple user interface for data collection, as shown in Figure~\ref{fig_interface}.
We showed a prompt image on the left panel of the interface and users {created} 
a corresponding sketch on the drawing canvas on the right.
To assist users in drawing proper proportions of the sketches, we implemented an \emph{optional} scaffold drawing mode in our interface besides the content drawing mode.
As shown in Figure \ref{fig_interface}, the scaffold lines (created in the same way as the content lines) were rendered in light blue and users could choose to hide/show the scaffolds while previewing/drawing the current sketch. 
Both the content strokes and the scaffolds were recorded in {separate} vector sequences to facilitate their independent analysis in the individual stroke level.
Users could change the stroke weights for their depiction.
Additionally, with digital settings, the stroke weights could also be anchored to the drawing pressure by checking the ``Pressure'' checkbox above the drawing canvas. \ac{Note that we recorded the pen pressure data regardless of whether the checkbox was checked.}
We also provided the undo/redo option and an eraser tool for modifying the currently drawn sketch. 
Our interface ran on Microsoft Surface Pro 4 with a Surface pen to mimic a paper-and-pencil setting for the convenience of {drawing}. 

\subsection{Participants}

We recruited two groups of in total 108 users to participate in our data collection task.
One group of 38 participants were experienced users with at least 5 years of professional drawing training/experience. Hereafter, we denote them as ``professional users''. 
The other group containing 70 users with less than 1 year of drawing experience, we denote them as ``novice users''.
The professional users aged from 18 to 38 and most of them were art school or design school students.

Compared to the professional group, the novice group had a more diverse background. The age range of the novice group was 18 to 46, and apart from the school students, the novice users also had more working experiences.
Each professional user drew 3-188 sketches (47.63 on average), and {each} novice user drew 3-137 sketches (25.86 on average). 
Among all the participants, only two (one professional and one novice) were reported as left-hand dominant. All the other users were right-hand dominant.
The demographic summary of the participants can be found in Table \ref{tab_demo}.

\begin{table}
\caption{Statistical summary of the participants in sketch collection. The values in the parentheses are the corresponding mean values.}
\label{tab_demo}
\begin{tabular}{c|c|c}
\hline
 & Professional & Novice \\
\hline
Gender & F: 28; M: 10 & F: 44; M: 26  \\
Age & 18-38 (26.61) & 18-46 (25.09) \\
{\#} Drawings & 3-188 (47.63) & 3-137 (25.86) \\
{\#} Sessions & 1-16 (4.47) & 1-9 (1.99) \\
\hline
\end{tabular}

\vspace{-8mm}
\end{table}

\subsection{Dataset}
After excluding the badly drawn sketches (in total 123 drawings \ac{almost without correspondences with the prompting shapes),} we made up for the removed sketches from more users 
and finally got a dataset containing 3,620 freehand sketches with the resolution of $800 \times 800$. 
Each prompt image was drawn 10 times: 5 times by professional users and 5 times by novices.
We got a sketch dataset containing samples from balanced skilled users.
Apart from the shape information recorded, we also collected the time information in the stroke level, as well as the drawing pressure for each stroke.
As discussed in \cite{Wang2021Tracing}, tracings provide precise representations of the shape information. We thus recruited another group of three users for tracing line drawings over the image prompts, resulting in a dataset containing 362 tracings for all the registration in the analysis stage. 
Each tracing was confirmed by presenting the resulting tracings to at least two users to avoid bias in our collection. When a user disagreed with a specific depiction, he/she directly modified the tracing and passed the modified tracing to additional users to check until no modification was needed, i.e., whether all the shape information of prompts was depicted in the tracings.
Please find the thumbnails of the collected sketches as well as the corresponding prompts in the supplemental material. 

\section{Multi-level Registration}
In this section, we first present our pixel-level registration method for our collected dataset and then introduce how to construct multi-level registration for analysis on entire sketches, strokes, and pixels. 

\label{sec_registration}
\subsection{Pixel-level Registration}

To spatially analyze the individual sketches, the first step is to find the correspondence between the freehand sketches and their prompts.
In our study, we resort to the sketch registration to find such correspondences.
We treat the corresponding tracing (covering all of the shape details) as an accurate shape representation of each prompt image, as validated by \cite{Wang2021Tracing}. 
Thus, for each prompt, we can register the ten freehand sketches to align with the prompt by maximizing the correlation between the sketch and the corresponding tracing.

\begin{figure}[htb]{
    \centering
    \includegraphics[width=\linewidth]{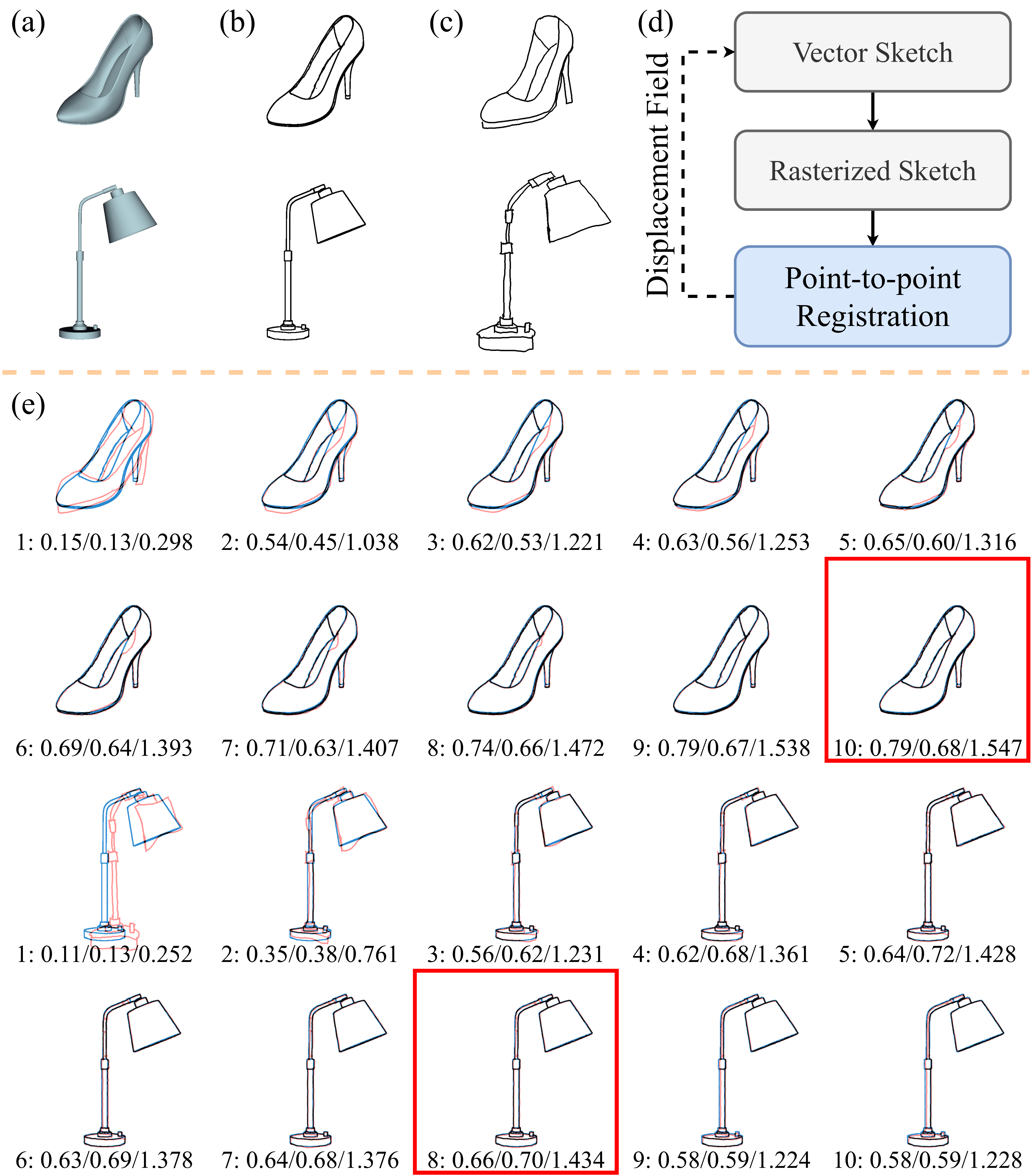}
    \caption{Examples of vector-sketch registration process. Given the model prompts in (a), (b) and (c) shows the corresponding tracings and users' drawings (from \textit{Novice 8} (Top) and \textit{Novice 18} (Bottom)).
    (d) illustrates the pipeline of our proposed pixel-level registration, which registers a sequence of vector strokes to the corresponding tracings by iteratively alternating rasterization and point-to-point registration.  
    (e) visualizes the registered results of each iteration, where the drawing pixels that have not been registered correctly are in red and the tracing pixels to be registered by the drawings are in blue, while the pixels overlapped by them are in black. The bottom text of each image indicates the performance evaluation (``$i$-th iteration: $E_i$ / $P_i$ / $R_i$''), by which our approach automatically picks the optimal iteration as the final result (in red bounding boxes). Please zoom-in to find the details of the registration progress. 
    }
    \label{fig:reg_progress}
    }
\end{figure}
Wang et al. \shortcite{Wang2021Tracing} adopted a point-to-point optimization-based registration method based on SimpleITK \cite{yaniv2018simpleitk} to register sketches to tracings. Although it works generally well on the sketches including texture and shading lines, it \ac{is not a good fit} for our collected sketches with sparse lines (only shape information). Specifically, the point-to-point registration might ignore the point order within strokes, making the vector sketches fail to align with the targets (i.e., the tracings), as shown in the top-left of each group in Figure \ref{fig:reg_progress} (e). Unlike the previous works \cite{cole2008people,Wang2021Tracing}, which mainly study where people draw given the shape prompts in a single level (i.e., at the pixel level), we investigate the drawings in a multi-level manner (Section \ref{sec:ml_geometry}), and thus need to maintain the order of points in a stroke intact in the analysis. 

To address the problem, we adopt an iterative rasterize-and-optimize process extended from the point-to-point registration method used in \cite{Wang2021Tracing}. See Figure \ref{fig:reg_progress} (d) for an illustration. Since the point-to-point registration easily \ac{stagnates} when there are only minor differences between the input and target due to the sparsity of sketches (as shown in the 4th and 5th iteration results of the shoes in Figure \ref{fig:reg_progress} (e)), we dynamically increase line width $l$ for re-rasterizing the vector data to gradually enlarge the registered difference. At the initial pass, the point-to-point registration between the rasterized vector sketch (represented as a binary image) and the corresponding tracing produces a displacement field at full resolution ($800 \times 800$), by which we update the positions for the points of the original vector sketch. Then we re-rasterize the new vector sequences and feed the newly rasterized image to the optimization process to predict a new displacement map. Using this iterative process, we get the well-registered vector sketches.

In our implementation, to handle drawings of different complexity levels, we run ten iterations of the above process in total (which is empirically enough to accurately align the vector sketches to the tracings). Then, we adopt the precision $P$ and recall $R$ to automatically pick the optimal registered result corresponding to the $i^*$-th iteration, and to avoid over-registration where different shape lines are merged together.
Specifically, it is formulated as follows: 
\begin{equation}
    i^* = \  \underset i{\mathrm{argmax}\;}E=\underset i{\mathrm{argmax}\;}\omega P_i+R_i, \ \ \ \ \mathrm{with}
    \label{eq:E}
\end{equation}
\begin{equation}
    P_i=\frac{o\_num_i}{reg\_num_i},\;R_i=\frac{o\_num_i}{trac\_num},
    \label{eq:PR}
\end{equation}
where $reg\_num$ and $trac\_num$ \ac{are respectively the numbers of filled pixels in} the registered result and the tracing, while $o\_num$ is the number of the overlapping pixels between them. For dynamic line width, we set $l=1$ for the first five iterations, and $l=i-4$ for the last five ones. Empirically, we set $\omega$ as $1.1$.  Figure \ref{fig:reg_progress} (e) shows the progressive registration results rasterized from the vector data \ac{by our method, where the first-iteration (top-left) results are equal to those generated by Wang et al.’s method.} Since this way actually registers the sketches point-by-point, we denote it as pixel-level registration. Please find more details and examples of pixel-level registration in \ac{Section 2 of Supplement}.

\begin{figure}[t]{
    \centering
    \includegraphics[width=\linewidth]{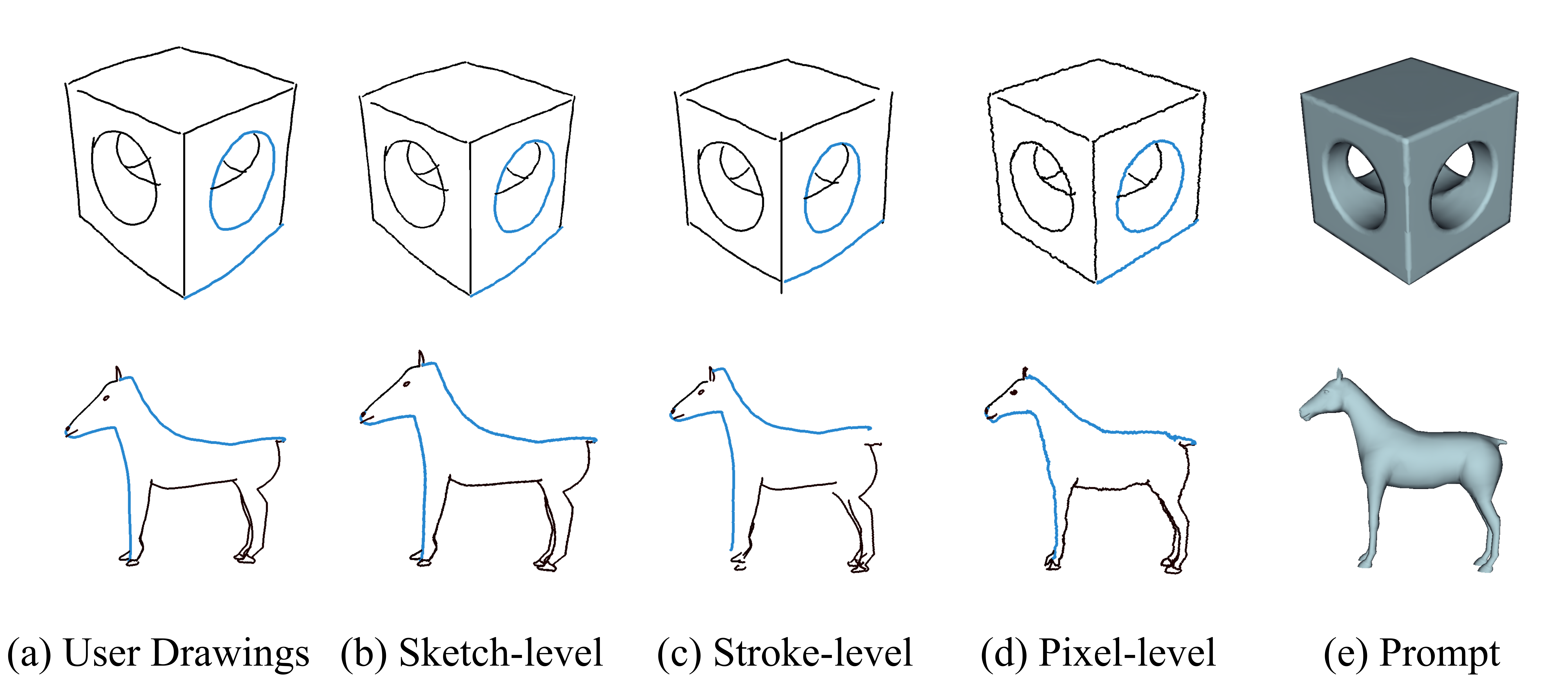}
    \caption{Multi-level transformation for analysis. The blue strokes are highlighted for {comparing their shape changes under different levels of registration}. 
    The two examples were drawn by \textit{Professional 2} and \textit{Novice 57}.}
    \label{fig:ml_geometry}
    }
\end{figure}

\subsection{Multi-level Scheme}
\label{sec:ml_geometry}
Prior works \cite{cole2008people,Wang2021Tracing} study human drawings registered in the pixel level only. Although the pixel-level registration (Figure \ref{fig:ml_geometry} (d)) is sufficient to investigate \emph{where} people draw the content, 
it generally washes away the drawing characteristics of sketches in the stroke or sketch level, thus hindering research aiming at exploring \emph{how} people draw the content and their performance variations, in the sketch and stroke levels. 
As shown in Figure \ref{fig:ml_geometry} (a) and (d), some strokes (in blue) lose their original geometry after pixel-level registration, e.g., the curvature changes in lines, the ellipse radii difference, the ellipse perspective variation, etc. 
This is mainly because the highly nonlinear deformation induced by pixel-level registration factors out shape variations in freehand sketches.

To preserve the sketch-level and stroke-level shape from users, we use two additional registration schemes. 
We estimate a similarity transformation (i.e., involving a rigid transformation and uniforming scaling) over the whole sketch or an individual stroke, respectively. 
Given the pixel-level registered results and the original vector data, we have two matched point sets ($dst\{i\}$ and $src\{i\}$) to be fitted by an optimal similarity transformation.
This can be formulated as the following minimization problem: 
\begin{equation}
    \lbrack {R_s}^\ast\vert T^\ast\rbrack=arg\underset{\lbrack R_s\vert T\rbrack}{min}\sum_i\parallel dst\lbrack i\rbrack-{R_s}src{\lbrack i\rbrack}
    ^T-T\parallel^2,
    \label{eq:warp_estimation}
\end{equation}
where $R_s$ indicates a matrix containing rotation $R$ and uniform scaling $S$ factors, and $T$ denotes a translation matrix. 
For sketch-level registration, we solve Equation \ref{eq:warp_estimation} by setting $dst\{i\}$ and $src\{i\}$ containing all the respective points in the original drawing and the corresponding pixel-level registered result for each sample, while for stroke-level registration, we process strokes individually by setting $dst\{i\}$ and $src\{i\}$ in each stroke for one drawing. 
After the application of the transformation matrix $[R_s|T]$ to {each drawing}, we get the multi-level registered sketches for the subsequent analysis.
As shown in Figure \ref{fig:ml_geometry}, a sketch-level transformation is the best to globally preserve user-drawn geometry and the pixel-level registration aligns the drawings almost perfectly fitting to
the target shapes (prompts), while stroke-level registration is a compromise between them. 

\section{Data Analysis}
\label{sec:analysis}
In this section, we provide the statistics and comparisons between the drawings by novice users and professional users from {both spatial and temporal} aspects. We also conducted an analysis on the similarity between the synthetic drawings and the freehand sketches, hoping to provide more inspirations for further design of drawing-generation methods.

\subsection{What Content Do People Draw?}
\label{sec:common}

\begin{figure}[htb]{
    \centering
    \hspace{-5pt}
    \subfloat[Common pixel distance]{
        \includegraphics[width=.495\linewidth]{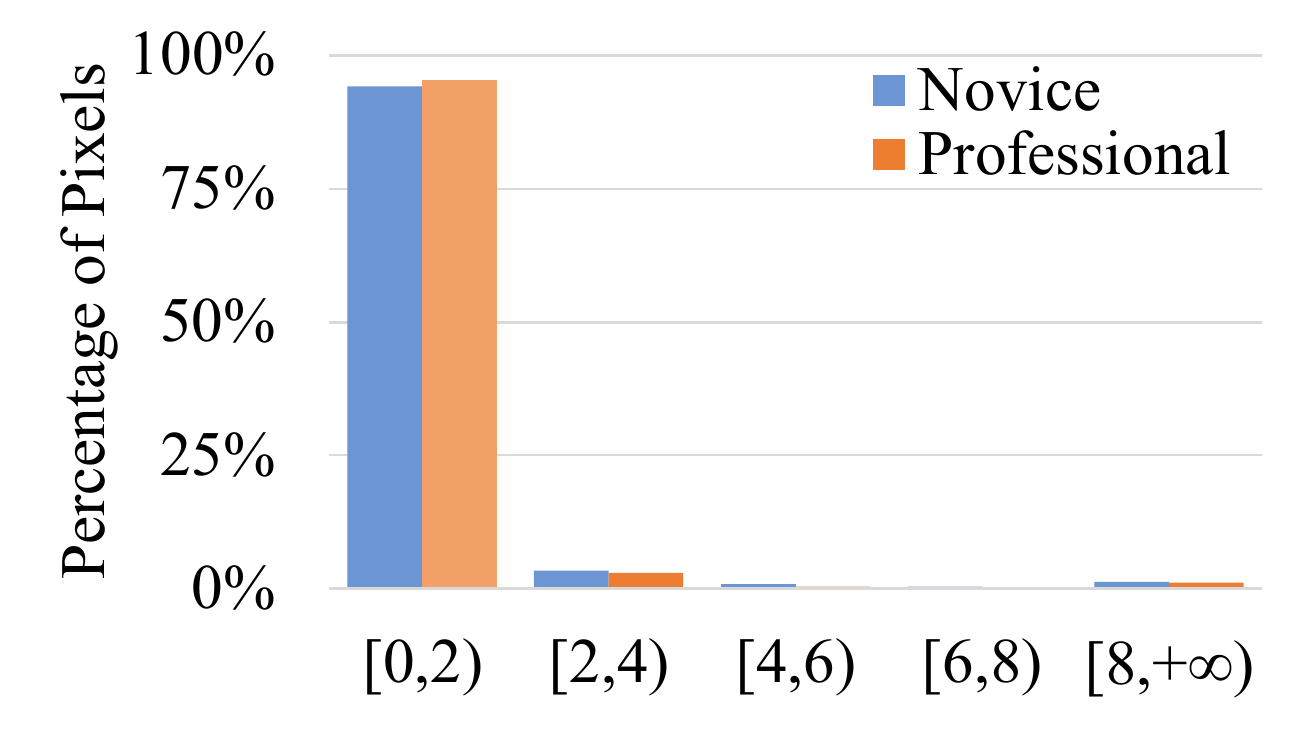}
    }
    \subfloat[Warped pixel displacement]{
        \includegraphics[width=.495\linewidth]{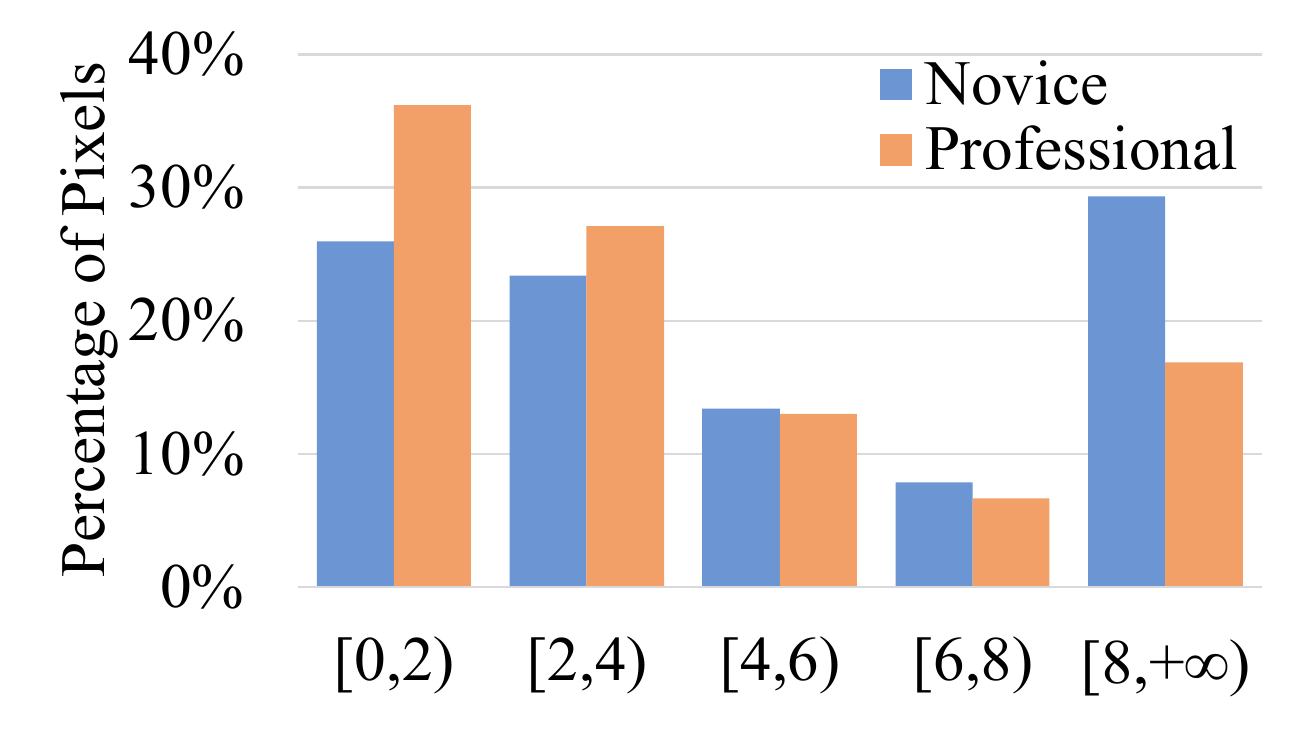}
    }\\
    \subfloat[Commonly drawn region (CDR)]{
        \includegraphics[width=\linewidth]{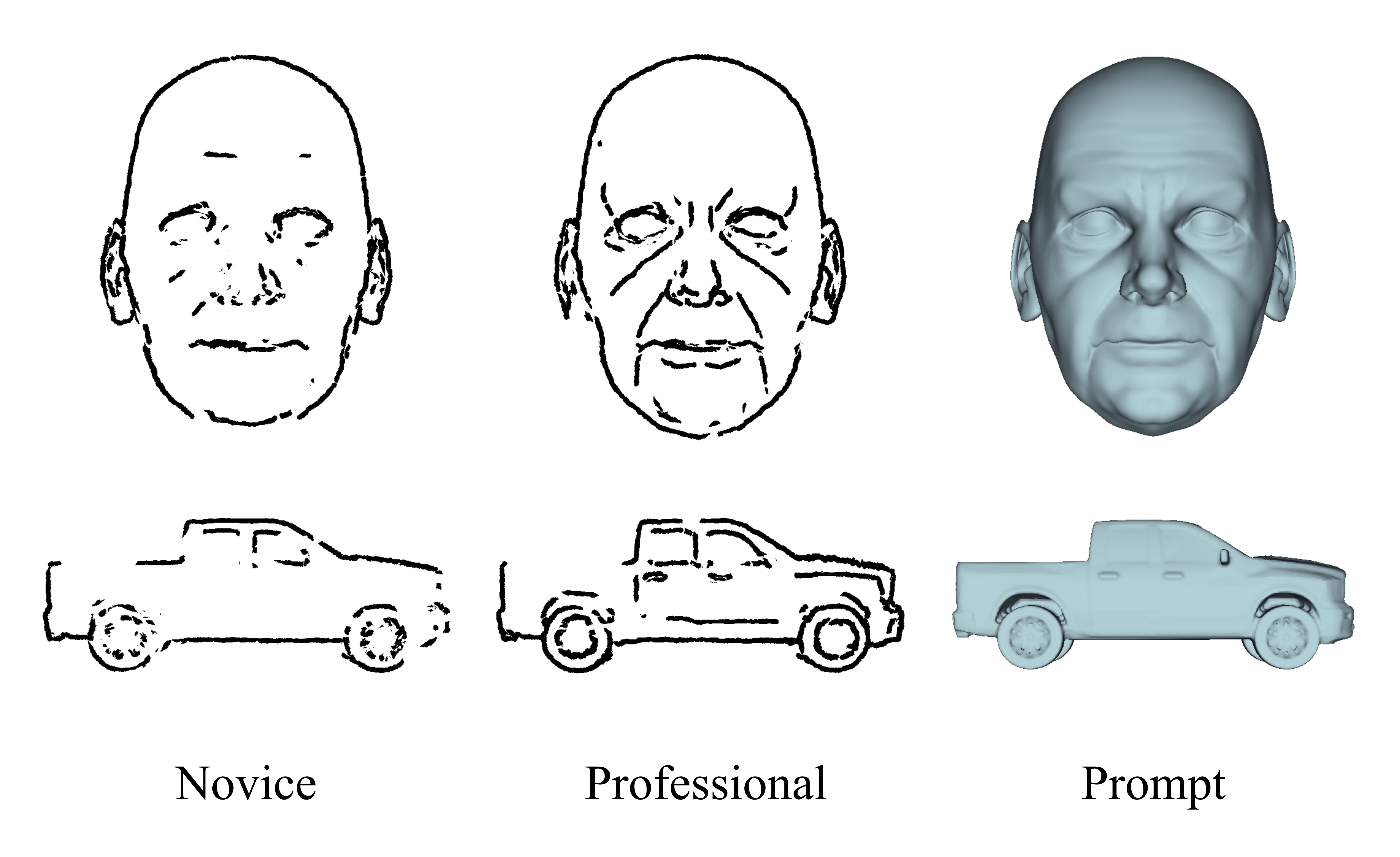}
    }
    \caption{Pixel-level similarity between novices' and experts' drawings. (a) \ac{Histograms} of pixel-wise closest distances from novice drawings to expert drawings (blue curve) and vice versa (orange curve) for all the categories. (b) \ac{Histograms} of the pixel distances between the stroke-level registered and pixel-level registered results, showing the local difference on each stroke (see Section \ref{sec:factors}). (c) CDR examples from novice and expert groups for the same prompts. 
    }
    \label{fig:pixel_stat}
    }
\end{figure}

Following the previous work \cite{cole2008people}, we first investigate the common content drawn by novice and {professional} users upon pixel-level geometry, i.e. rasterizing the registered vector sketches using one pixel-wide strokes. {Note that we only compute the valid drawings that can be correctly registered to the target, i.e., $E^*$>1.2 (Eq. \ref{eq:E}).}
To compare the individual drawings (after pixel-level registration) from the two groups, for each pixel in each drawing from the novice group, we find the closest pixels in all of the drawings from the professional group for the same prompt and compute their distance in pixel, and vice versa (i.e., from the drawings in the professional group to the drawings in the novice group).
We plot \ac{histograms} of the pixel-wise distances across all the categories respectively for the two groups.
As shown in Figure \ref{fig:pixel_stat} (a), the two \ac{histograms are similar to each other.}
We also visualize common contents from the two groups by superimposing the registered results (Figure \ref{fig:teaser} (c)).
It reflects that for individual drawings, not only professional users drew similar contents among themselves \cite{cole2008people}, but also novice users drew contents which {professional} users tend to draw.

To find out how consistently the two groups drew the contents, inspired by \cite{Wang2021Tracing}, we extracted a binary commonly drawn region (CDR) \ac{drawn by all the users in each group.}
Similarly, we obtained a distribution of pixel-wise distances between CDR pixels for the novice and professional groups. 
The main gap between the two groups falls into the categories containing detailed contents, e.g., \textit{Human Face}, \textit{Vehicle}, and \textit{Shoes}. {Professional} users have higher consistency in drawing detailed parts of 3D models than novices, though the two groups would commonly draw general shapes of objects, including obvious silhouettes and occluding contours, as shown in Figure \ref{fig:pixel_stat} (c). The tendency is also reflected in the basic statistical analysis \ac{(see the differences of stroke count and path length in Section 3.1 of Supplement)}.

\textbf{Conclusion}: Both the novice and professional users perceived almost the same shape features from 3D models, but the professional users tended to draw more details given more complicated 3D models.

\subsection{Where Does Key Difference Occur When Sketching?}
\label{sec:factors}

Although novice and professional users would draw similar contents existing in 3D objects, \ac{where does the key difference occur}? To understand how differently they sketch a given prompt, we analyze the errors between their original drawings and the prompts from three aspects (sketch-level, stroke-level, and pixel-level), with the help of multi-level registrations.

For qualitative comparisons, Figure \ref{fig:teaser} shows three groups of paired drawings by superimposing five drawings respectively from the novice and professional users for the same prompts after multi-level registrations. We found that sketch-level registered drawings by professionals were
already very close to the 3D shapes both globally and locally. In contrast,  
those by novices are messy in local parts, though their main structures can still be recognized to be the prompting categories. It indicates that novice users could not maintain well part-by-part ratios among each other following the prompts. Compared between the next two levels (Figure \ref{fig:teaser} (b) and (c)), there is a larger improvement existing in the novice drawings from stroke-level to pixel-level registration. It means that unlike professional drawings, novice drawings require more nonlinear transformations for each stroke to change the original geometry drawn by novice users to match the targets. We will show quantitative comparisons below to support these findings. 
Note that for the sketch-level comparison, we only count the valid drawings like Sec. \ref{sec:common}, while for the stroke-level and pixel-level ones, we only count the valid strokes that can be correctly registered to the tracings, i.e., the overlapping rate is higher than 80\%.

\begin{figure}[t]{
    \centering
    \subfloat[Sketch-level]{
        \begin{minipage}{.49\linewidth}
          \includegraphics[width=\linewidth]{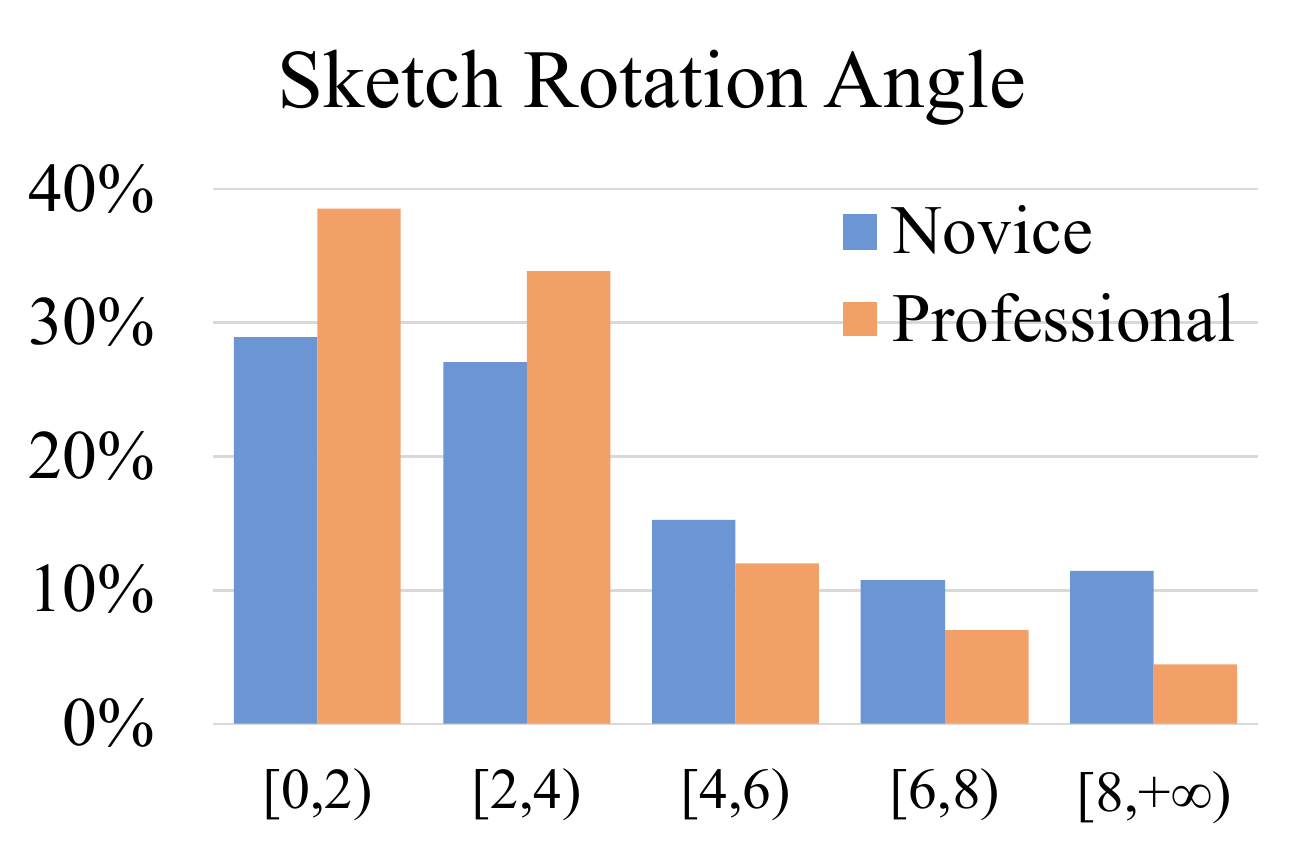}
          \includegraphics[width=\linewidth]{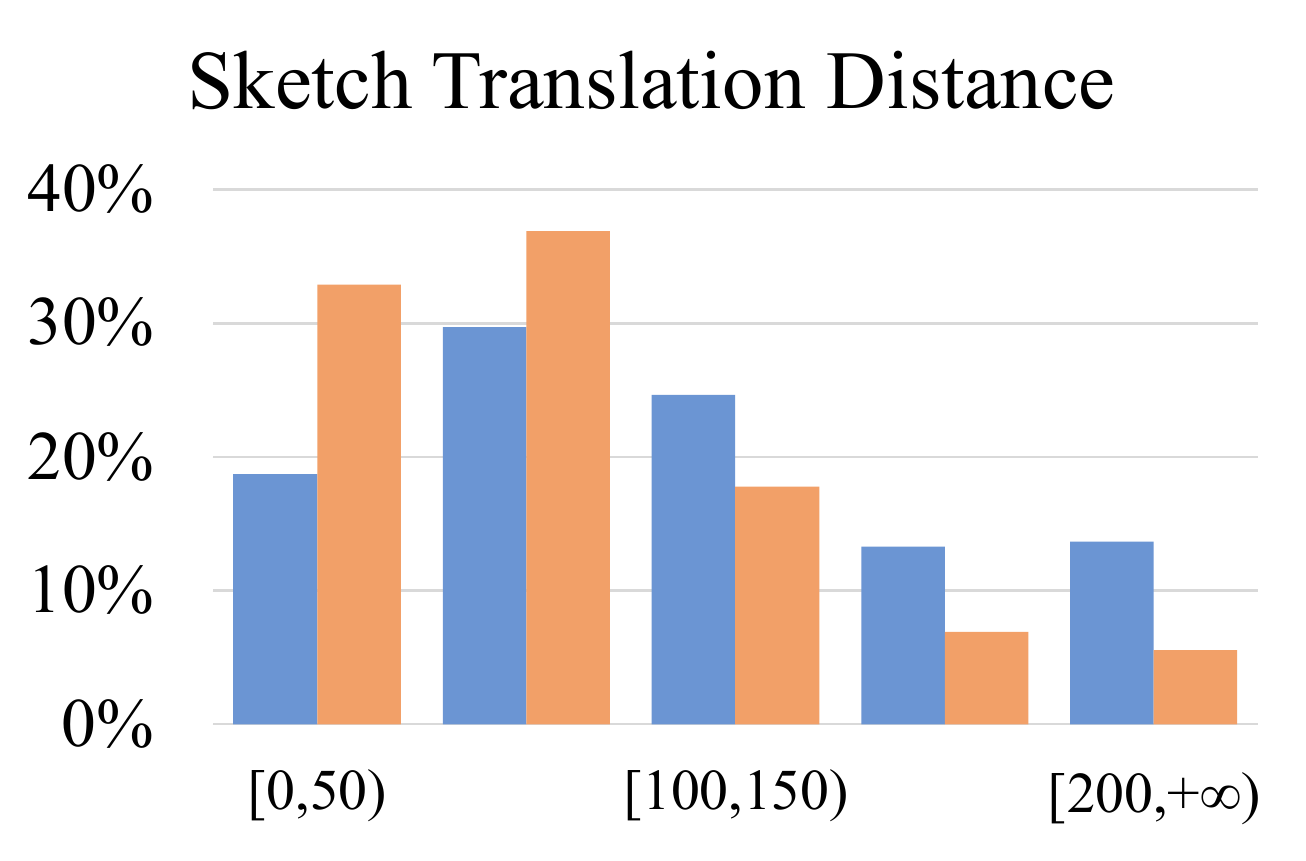}
          \includegraphics[width=\linewidth]{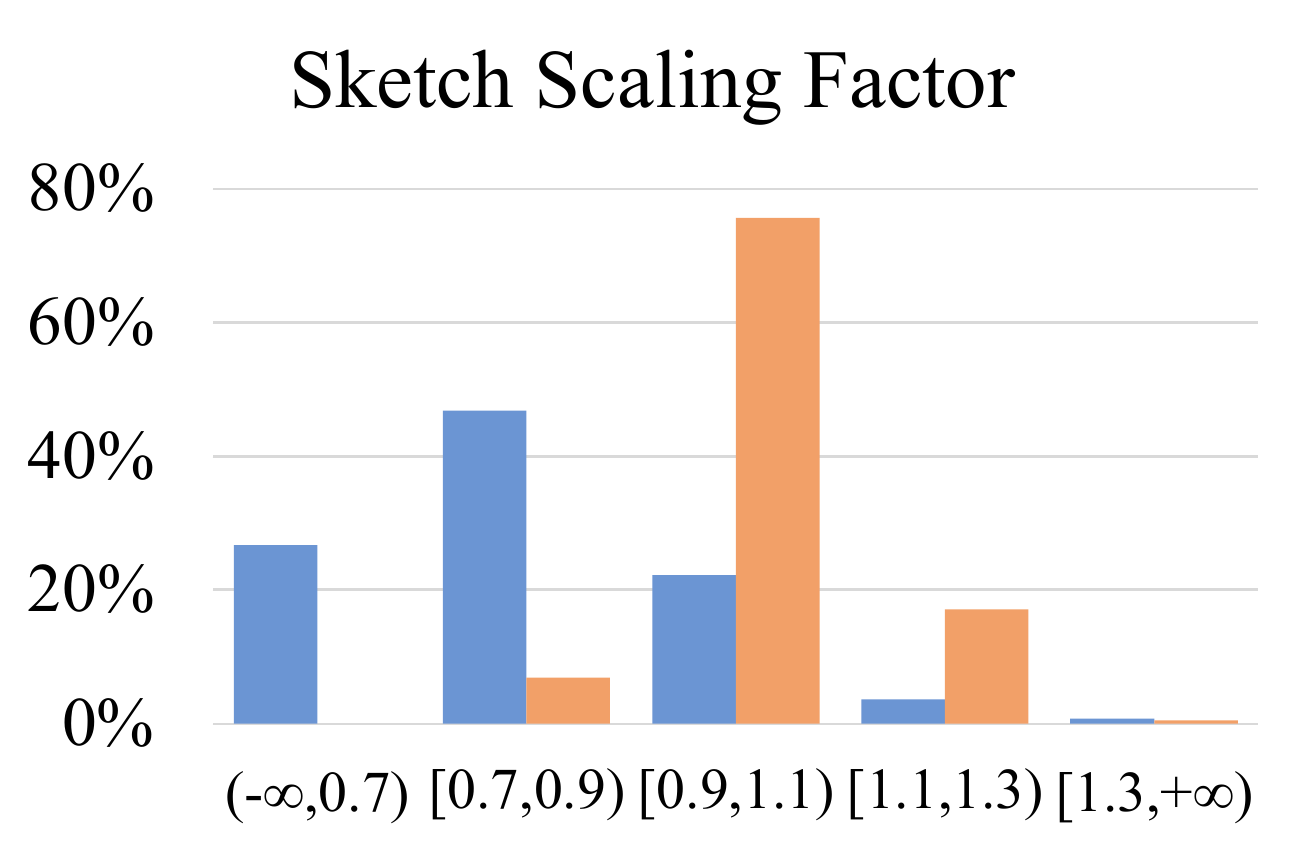}
        \end{minipage}
    }
    \subfloat[Stroke-level]{
        \begin{minipage}{.49\linewidth}
          \includegraphics[width=\linewidth]{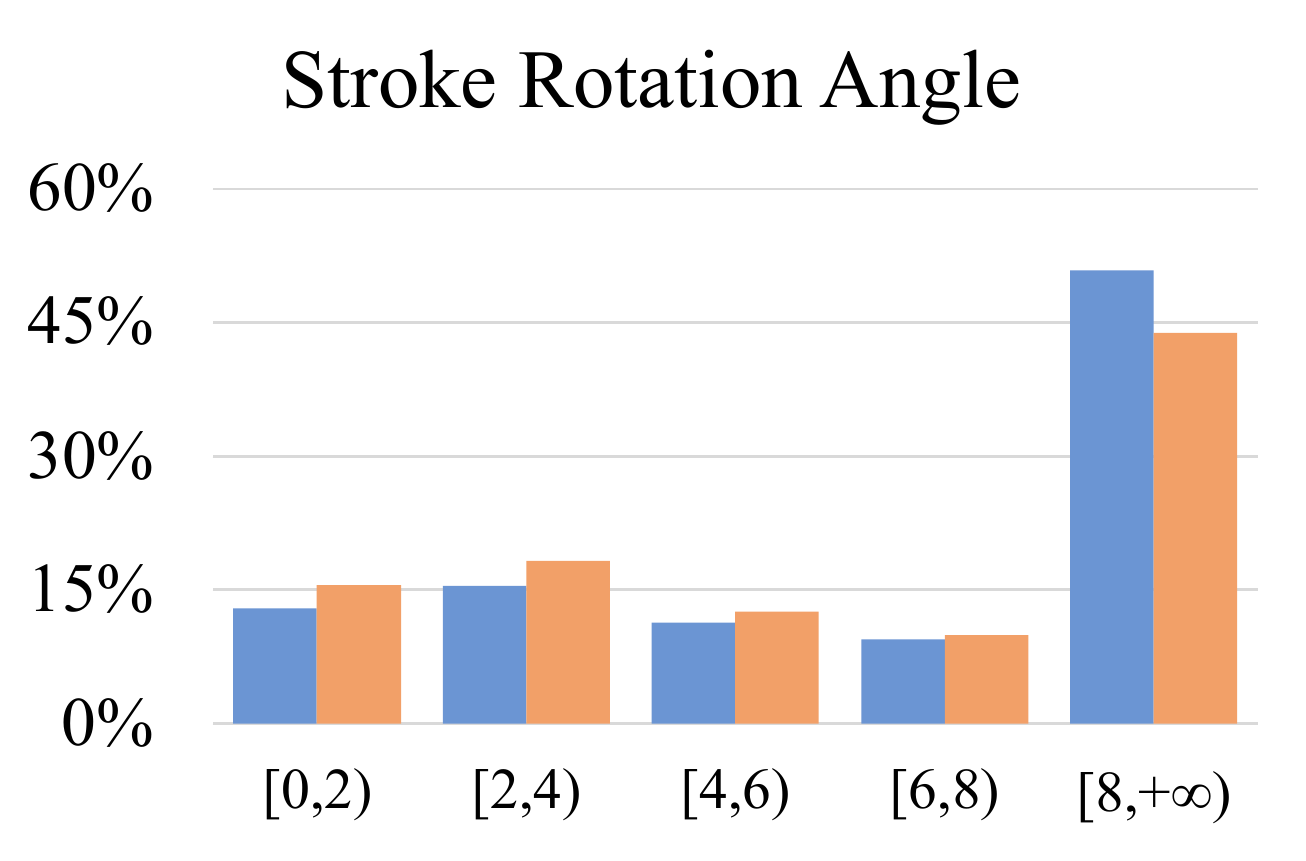}
          \includegraphics[width=\linewidth]{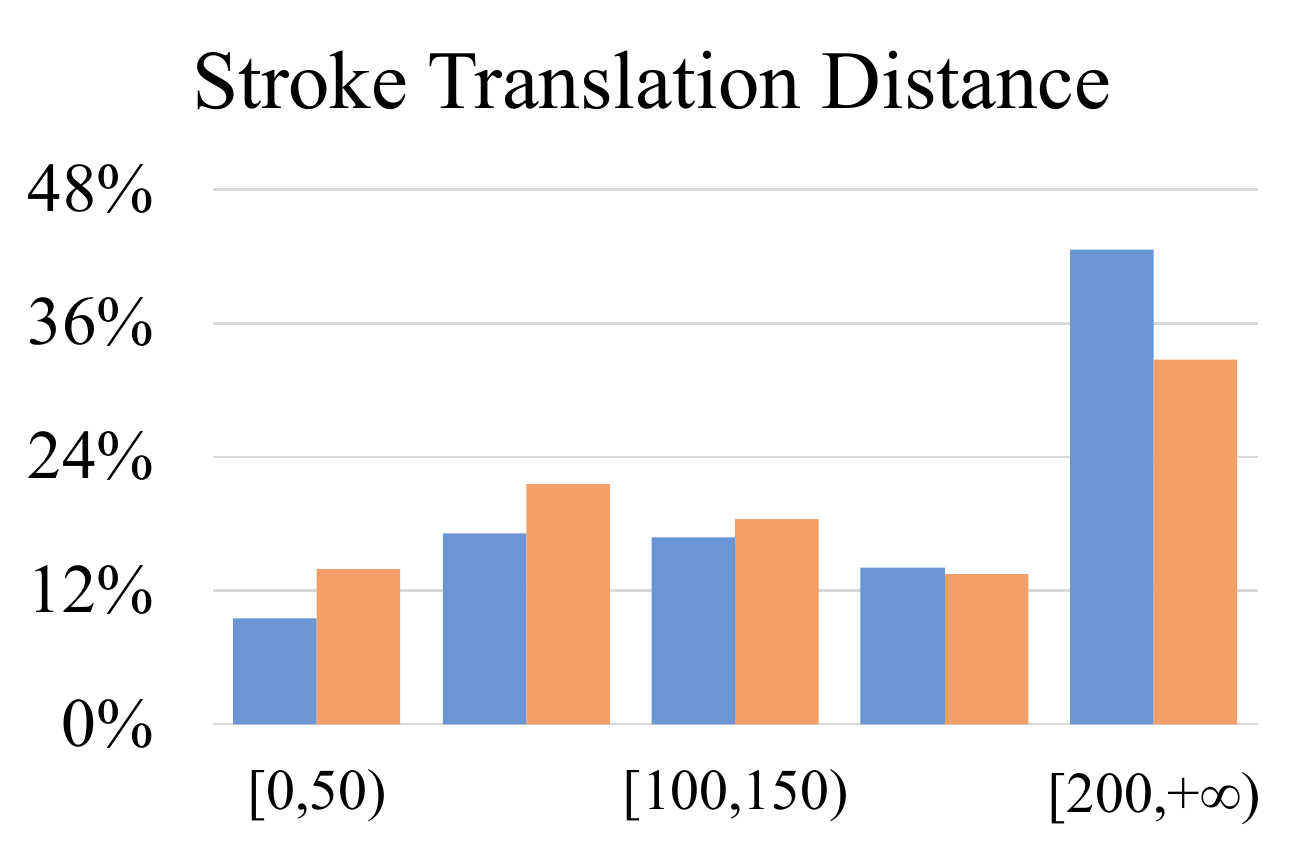}
          \includegraphics[width=\linewidth]{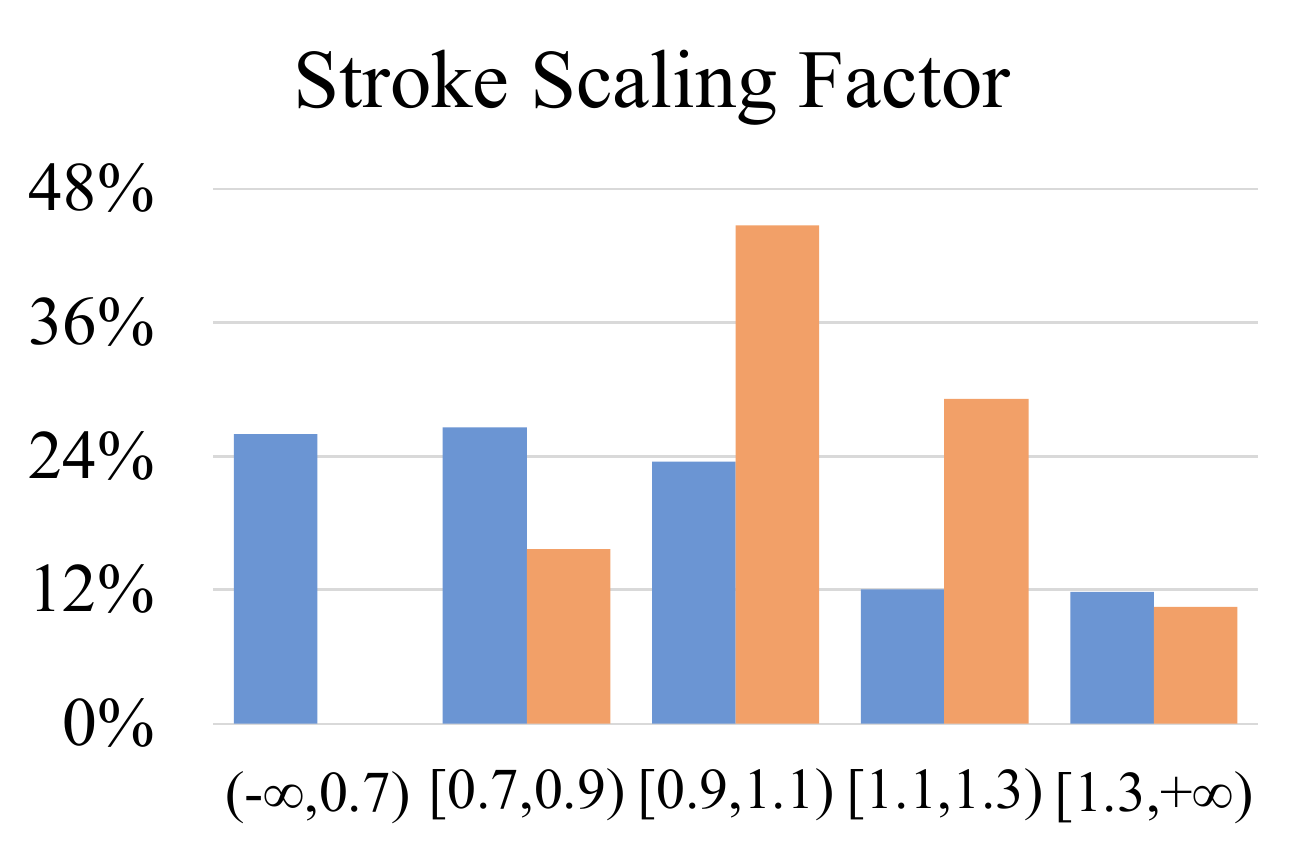}
        \end{minipage}
    }
    \caption{{\ac{Histograms} of {the} three parameters of similarity transformation for sketch-level (a) and stroke-level (b) for novice group and professional group across all the categories. Note that for stroke-level evaluation, the parameters are relative to those of sketch-level transformation.}}
    \label{fig:stroke_stat}
    }
\end{figure}

From a sketch-level perspective, we explore how people organize their sketches globally 
by measuring the difference between the original drawings and the corresponding results after the sketch-level registration.  
We compute three \ac{histograms} 
across all the categories for the novice and professional groups, respectively, including rotation angle ($R_G$), translation distance ($T_G$), scaling factor ($S_G$), as shown in Figure \ref{fig:stroke_stat} (a). 
Note that we compute the sketch-level translation to show the users’ habit of placing a drawing relative to an object center, but not to reflect the geometry error. 
To observe the difference \ac{on organizing} individual strokes, we also evaluated the relative transformation parameters ($R_L$, $T_L$, $S_L$) for each stroke between sketch-level and stroke-level registration, i.e. the local error relative to the global error. Similarly, we plotted three \ac{histograms}
on the strokes across all the drawings for the novice and professional groups (Figure \ref{fig:stroke_stat} (b)). 
To further evaluate the difference of stroke geometry between the two groups, we measure the distance between stroke-level registered results and pixel-level registered results for each drawing point, i.e. pixel-level translation, since pixel-level registration can produce the results highly aligned with the prompts. Figure \ref{fig:pixel_stat} (b) shows the \ac{histograms} of the novice and professional drawings. \ac{Please find more comparisons in Section 3.2 of Supplement.}

\textbf{Conclusion}: The novice and professional users had similar senses of understanding the global orientation of the prompting 3D models and tended to follow the geometry when sketching globally. The key difference between them is \ac{that} the professional users performed significantly better on preserving \ac{the global layouts of the sketches as well as the extrinsic and intrinsic properties of the strokes, especially for scaling factors, to match with the prompts well.}

\ac{
\subsection{Could Scaffold Lines Help Improve Sketches?}
\label{sec:scaffold}
Gryaditskaya et al.~\shortcite{gryaditskaya2019opensketch} show that the presence of scaffold lines has positive correlation with the accuracy of the professional sketches. Since we provided an option of scaffold drawing for users, we investigate whether scaffold lines could effectively help both novices and professionals improve their sketches and in which aspect the scaffold lines mainly contribute to. In our dataset, we found that more professional drawings (53.5\%) were completed with the help of scaffold lines, compared to novice drawings (36.7\%). Then, we evaluate the correlation between the accuracy of the two-group drawings and the usage of scaffold lines from global to local, i.e., sketch-level, stroke-level, and pixel-level.

For sketch-level and stroke-level, we utilize the respective transformation parameters ($R_G$, $T_G$, $S_G$ and $R_L$, $T_L$, $S_L$) (Section \ref{sec:factors}) to compute the drawing errors, including rotation error $E_{R}$=$R$ ($R$ indicating $R_G$ or $R_L$), translation error $E_{T}$=$T$, and scaling error $E_S$=$|S-1|$. 
For the pixel-level errors, we first compute the proportion of overlapping pixels between the stroke-level registered results and the corresponding tracings. 
Here, one stroke would be regarded as an incorrect drawing path if less than 50\% its pixels 
are overlapped. We then evaluate the pixel-level inaccuracy $E_P$ of a drawing through the ratio of the incorrect strokes. 
We separately measure the above errors on the novice and professional drawings without and with the scaffold lines and conduct statistical tests using a linear mixed model (LMM) with the R-style formula, e.g., ``$T\sim$ Condition + (1 | PromptID) + (1 | UserID)" for $T$, to show the significance of the difference. 
Table \ref{tab:scaffold} presents the average errors for the four conditions and the p-values of LMMs for the two groups. Please find more analysis in Section 3.3 of Supplement.

\textbf{Conclusion}: It is intuitive that scaffold lines effectively helped both the novice and professional users set up the global scale of the drawings as well as the local drawing paths (pixel-level) to approximate the prompt targets. It similarly provided useful guidance for the novices to place each stroke on its correct {position}, but contributed little on guiding orientation neither globally or locally.
In addition, scaffolding worked less significantly for the professionals since their sufficient drawing skills could compensate the case without scaffold lines.
\begin{table}[htb]
\small
\caption{The evaluated average errors and p-values of LMMs for the novice and professional drawings respectively without and with scaffold lines on sketch-level, stroke-level, and pixel-level. Each bold p-value indicates that there is a significant difference ($p$<0.001) between the conditions.
Note that we compute $E_{GT}$ to show the users’ habit of placing a drawing relative to an object center, but not to reflect the geometry error.
}
\label{tab:scaffold}
\begin{tabular}{c|ccc|ccc}
\hline
           & \multicolumn{3}{c|}{Noivce}        & \multicolumn{3}{c}{Professional}   \\ \hline
Error & w/o    & w/     & $p$-value          & w/o    & w/     & $p$-value           \\ \hline
$E_{GR}$   & 5.14   & 4.68   & 0.02            & 3.58   & 3.22   & 0.006             \\
$E_{GT}$   & 130.79 & 104.06 & <0.001 & 97.98  & 76.83  & <0.001 \\
$E_{GS}$   & 0.18   & 0.14   & \textbf{<0.001} & 0.12   & 0.09   & \textbf{<0.001}  \\ \hline
$E_{LR}$   & 13.51  & 13.59  & 0.43             & 11.28  & 11.06  & 0.003  \\
$E_{LT}$   & 242.01 & 245.80 & \textbf{<0.001}  & 199.04 & 196.63 & 0.004            \\
$E_{LS}$   & 0.29   & 0.28   & 0.02    & 0.23   & 0.23   & 0.01              \\ \hline
$E_P$  & 0.41   & 0.36   & \textbf{<0.001}  & 0.31   & 0.27   & \textbf{<0.001}  \\ \hline
\end{tabular}

\end{table}
}

\subsection{Do People Sketch 3D Shape Differently over Time?}

\begin{figure}[t]{
    \centering
    \includegraphics[width=\linewidth]{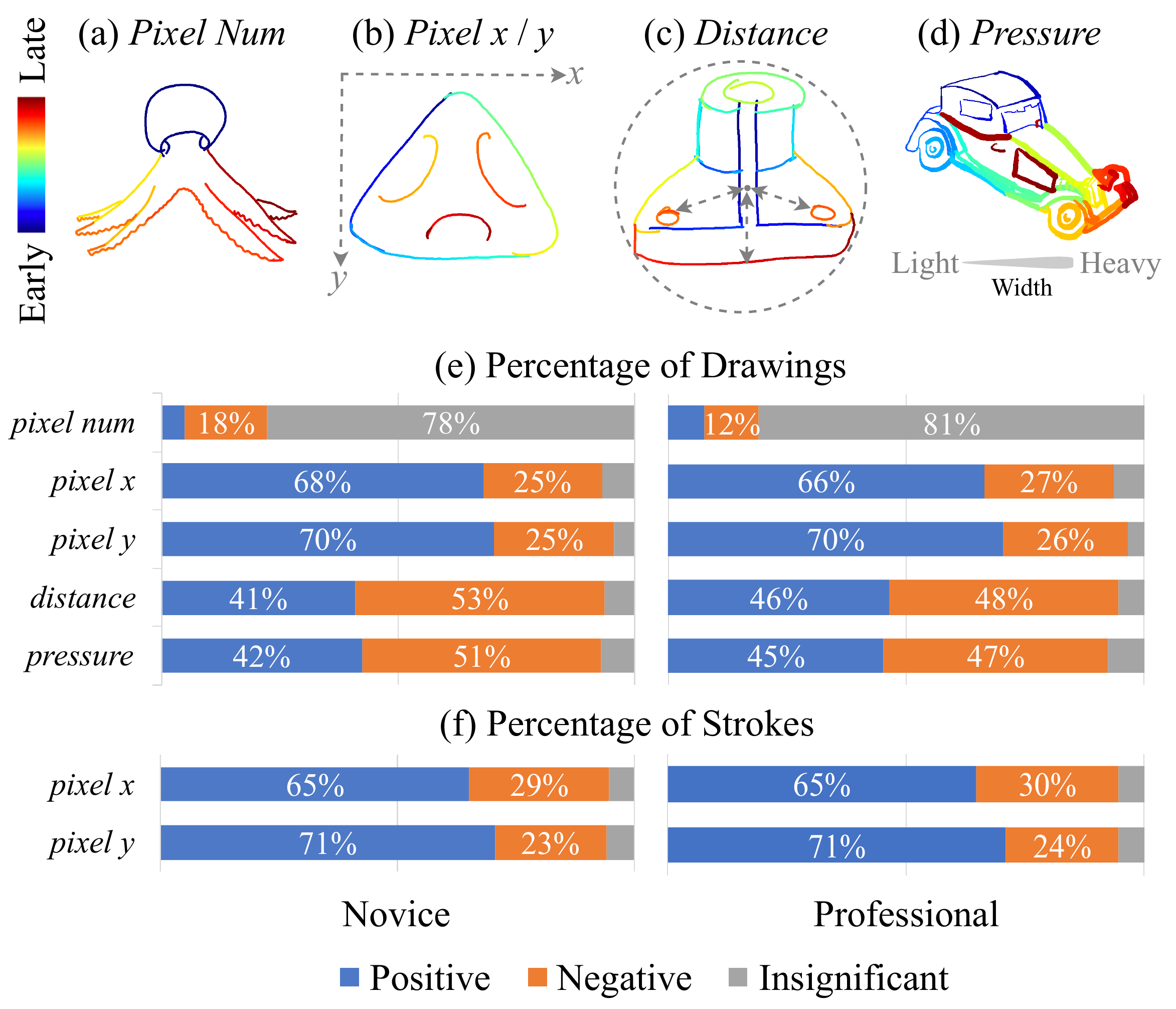}
    \caption{
    \ac{(a)-(d) Examples of drawings have positive correlations with time, respectively on pixel counts in even temporal bins, pixel locations ($x$,$y$), pixel distance to the drawing center, and pen pressure.}
    (e) \ac{Stacked bar charts} of Spearman correlation ($p$<0.001) for different drawing features over time on the two-group drawings. (f) \ac{Stacked bar charts} of Spearman correlation ($p$<0.001) on drawing each stroke.
    }
    \label{fig:temporal}
    }
\end{figure}

After evaluating the spatial differences between the two groups, now we evaluate their temporal differences. Similar to \cite{Wang2021Tracing}, we interpolated the timestamps for each pixel and assigned them into even temporal bins (25 bins) for each drawing. To observe the relationship between the number of human-drawn pixels and time, we computed Spearman correlation coefficient with a significance threshold $p$<0.001 between them, respectively for the novice and professional groups (1,810 drawings in each group). We found in most of the drawings (more than 78\%), the number of drawn pixel had no significant correlation with time. To study the relationship between spatial drawing location and time, we tested Spearman correlation coefficient on the pixel locations of raw user drawings over time, i.e., the coordinate $(x,y)$ and the distance to the drawing center. 
\ac{Figure \ref{fig:temporal} (a)-(d) present the examples with positive correlations between the features and time. Figure \ref{fig:temporal} (e) shows that both of the novices and professionals tended to draw pixels from left to right and from top to bottom on the canvas in most of the drawings (over 66\%).
Almost 50\% of the drawings had positive or negative correlations between the distance to the center and time. It means that people tended to either draw from outside to inside or from inside to outside, depending on the structure of a target object.}
We further evaluated how people drew each stroke spatially $(x,y)$ over time, and found the similar tendency with the case they drew the whole sketches (Figure \ref{fig:temporal} (f)). It is natural that the right-handed users (99\% in our user study) have such preferred drawing directions, as the stroke direction guideline documented by Fu et al. \shortcite{fu2011animated}. 
\ac{Please find more comparisons in Section 3.4 of Supplement. 
}

To evaluate the difference of stroke ordering between novices and professionals, we examine the raw user drawings based on the ordering guidelines \cite{fu2011animated}, including: 

\ac{
\textit{Simplicity}. People first draw lines with simple shapes or smooth lines, and then more complex curves.

\textit{Proximity}. People start the next stroke to the nearest element or the element along the trajectory they are following.

\textit{Collinearity}. People follow the trajectory of interrupted lines for continuity.

\textit{Anchoring}. People draw substrate strokes earlier than attachment strokes.
}

\begin{figure}[t]{
    \centering
    \includegraphics[width=.8\linewidth]{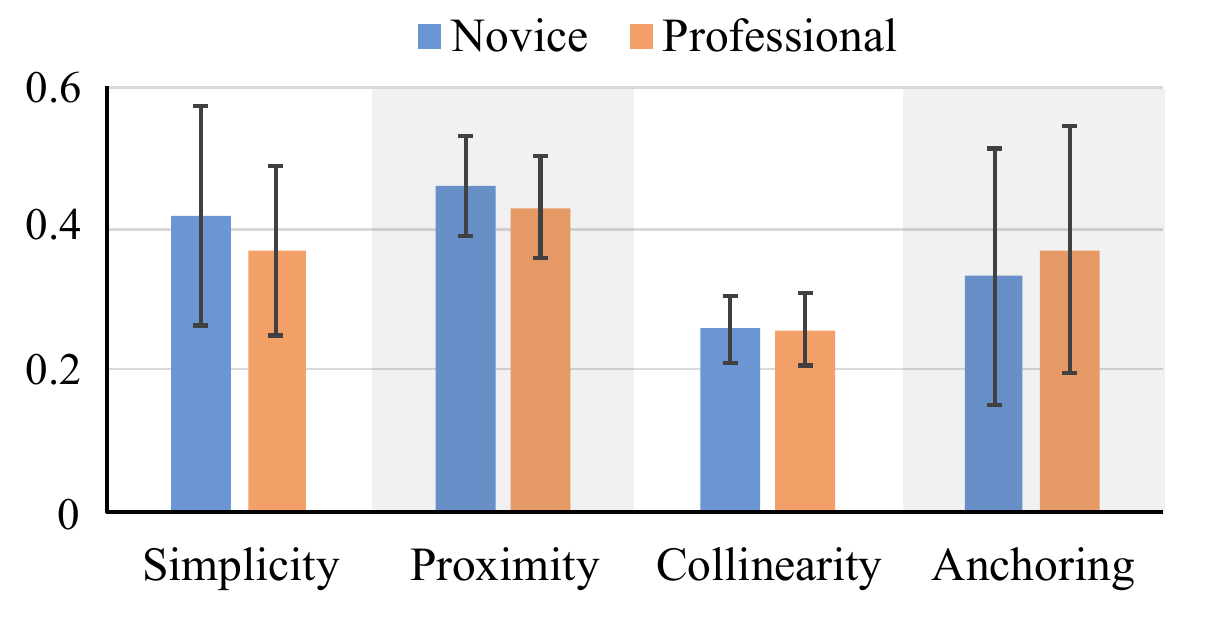}
    \caption{The mean and standard deviation of the energy costs for different ordering guidelines over the two-group drawings. The lower values means one group is guided more by a certain concerned 
    guideline than the other group.}
    \label{fig:guidelines}
    }
\end{figure}

We computed the energy costs of the guidelines for each drawing, following \ac{Fu et al. \shortcite{fu2011animated}  (see the definition in Section 3.4 of our Supplement).}
Figure \ref{fig:guidelines} shows the average cost of the human-drawn ordering over all the drawings from the novice or professional users for each of the above guidelines. It indicates that the professionals followed more on the simplicity and proximity guidelines than novices, while the novices followed more on the anchoring guideline. The two groups had similar ordering tendency on the collinearity. \ac{The $p$-values of LMMs (as done in Section \ref{sec:scaffold}) between the two groups also show that there are significant differences ($p$<0.001) on the simplicity, proximity, and anchoring guidelines, but not on the collinearity ($p$=0.006).}

\textbf{Conclusion}: Both of the novice and professional users similarly tended to follow the drawing direction guideline, i.e., from left to right and from up to down over time. Also, the simplicity and proximity guidelines \cite{fu2011animated} provided more guidance for the professionals than the novices, while the anchoring guideline worked more for the novices.

\begin{figure*}[htb]{
    \centering
    \subfloat[{Qualitative Results}]{
    \includegraphics[width=\linewidth]{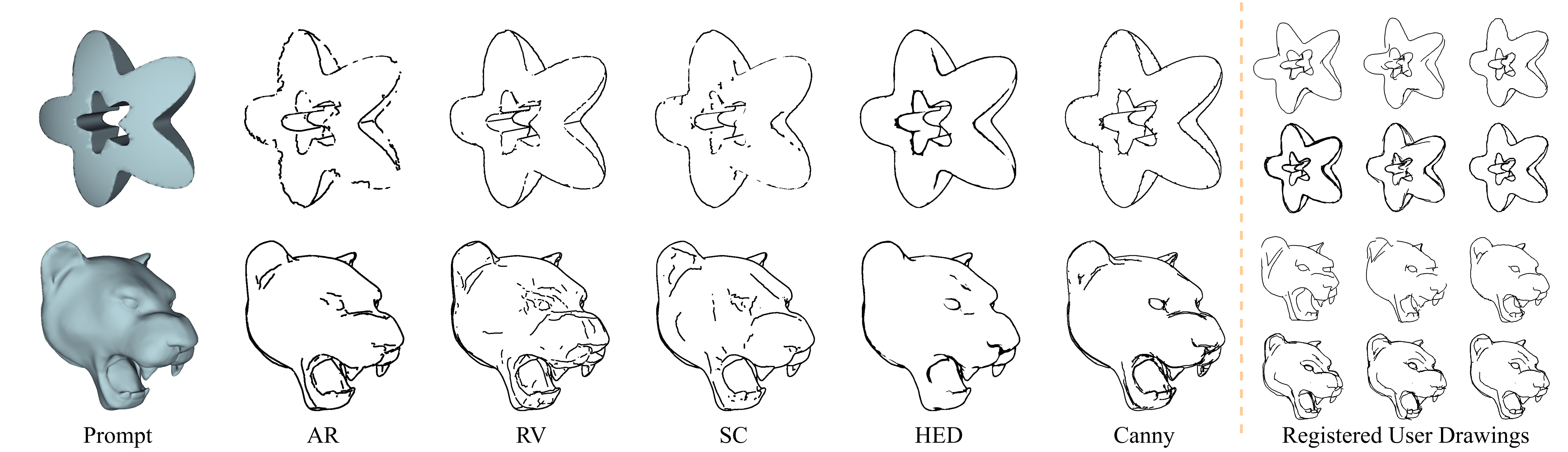}
    }\\
    \subfloat[Sketch-level]{
        \begin{minipage}{.3\linewidth}
          \includegraphics[width=\linewidth]{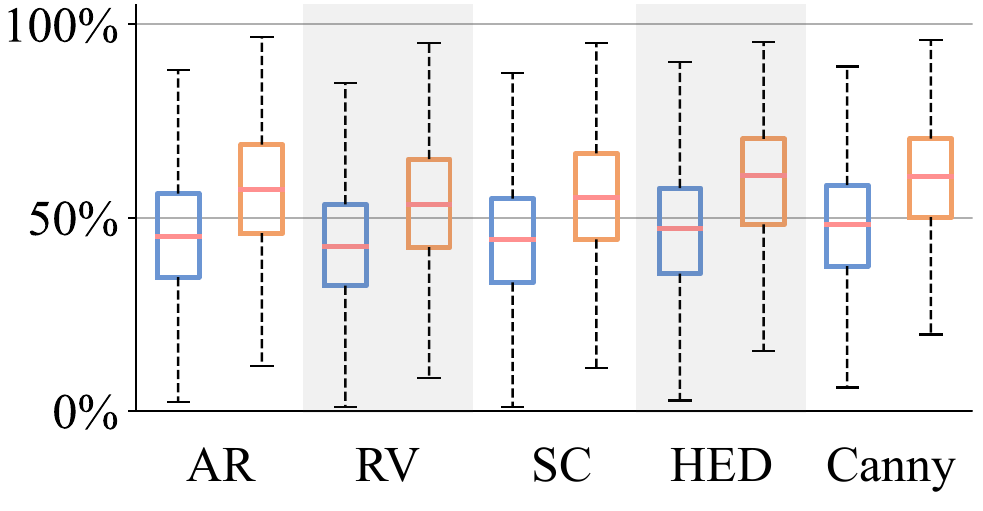}
          \includegraphics[width=\linewidth]{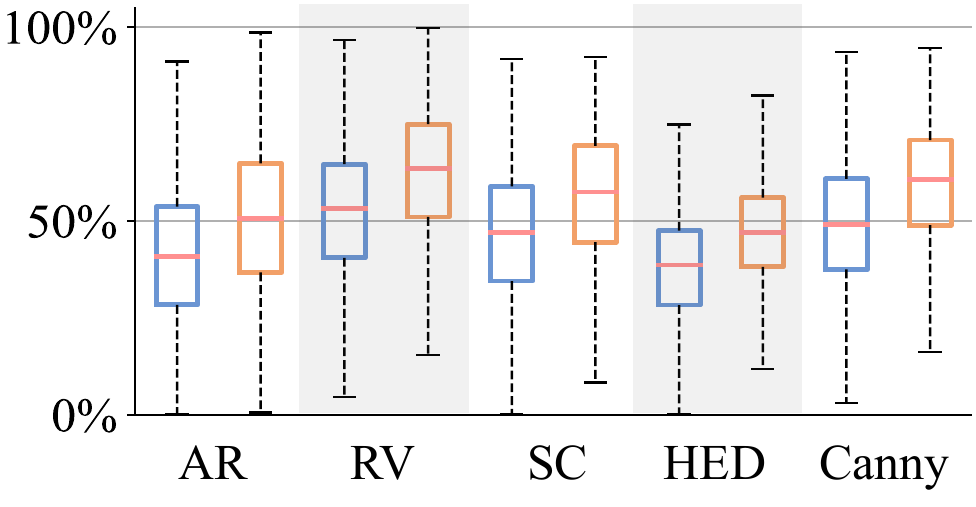}
        \end{minipage}
    }
    \subfloat[Stroke-level]{
      \begin{minipage}{.3\linewidth}
          \includegraphics[width=\linewidth]{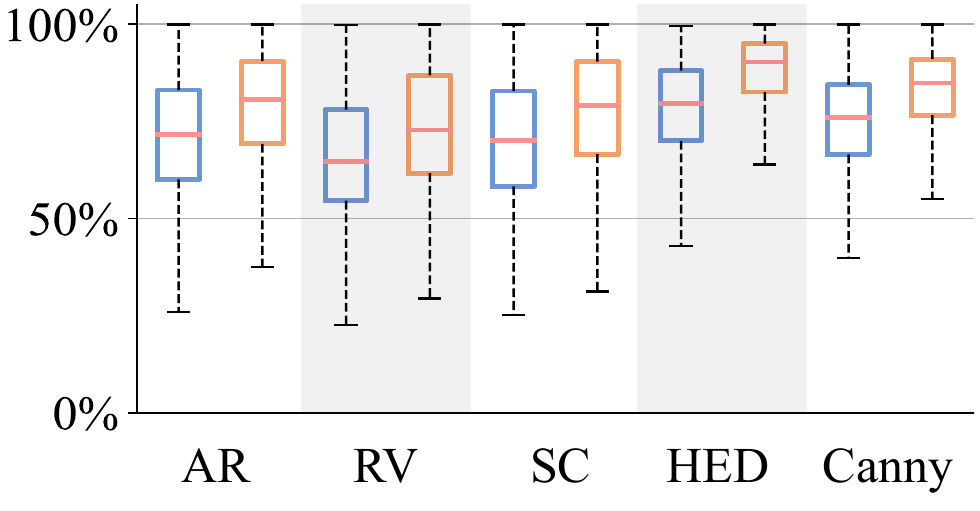}
          \includegraphics[width=\linewidth]{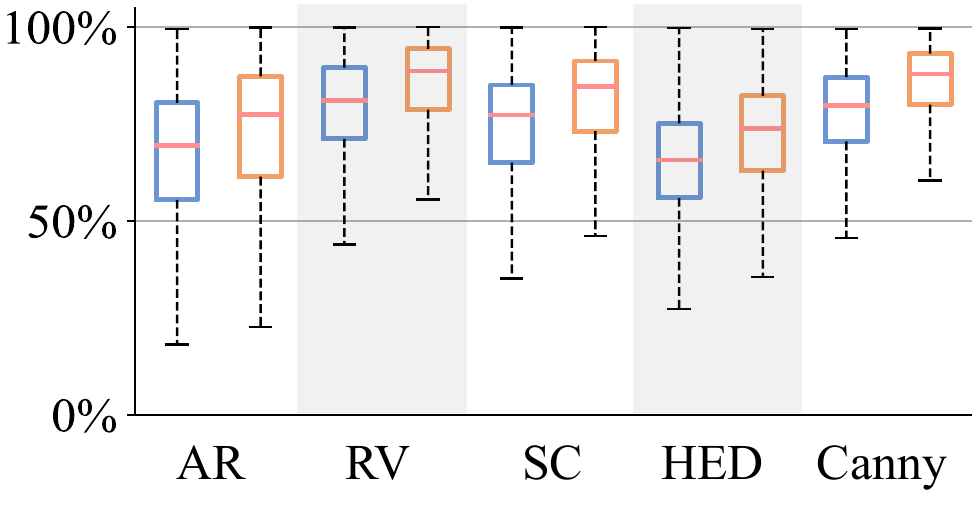}
        \end{minipage}
    }
    \subfloat[Pixel-level]{
      \begin{minipage}{.3\linewidth}
          \includegraphics[width=\linewidth]{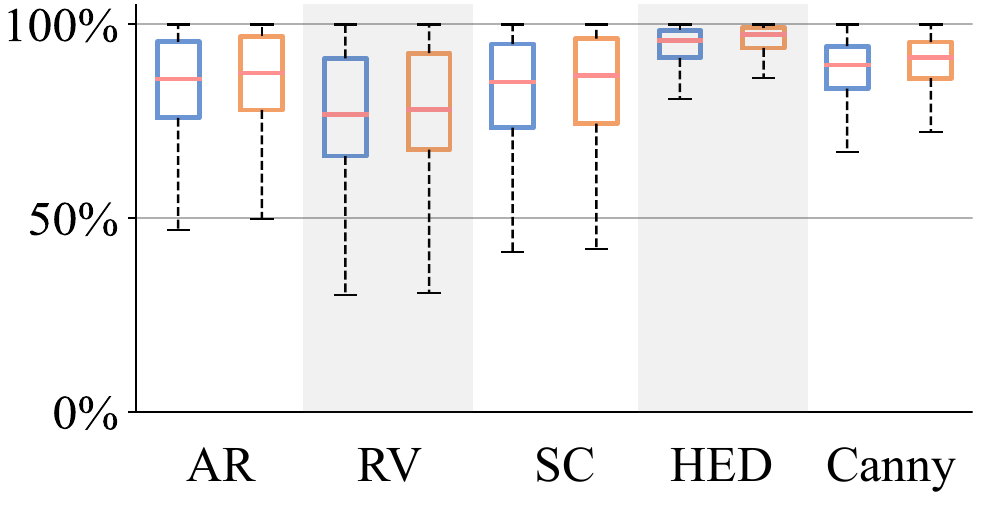}
          \includegraphics[width=\linewidth]{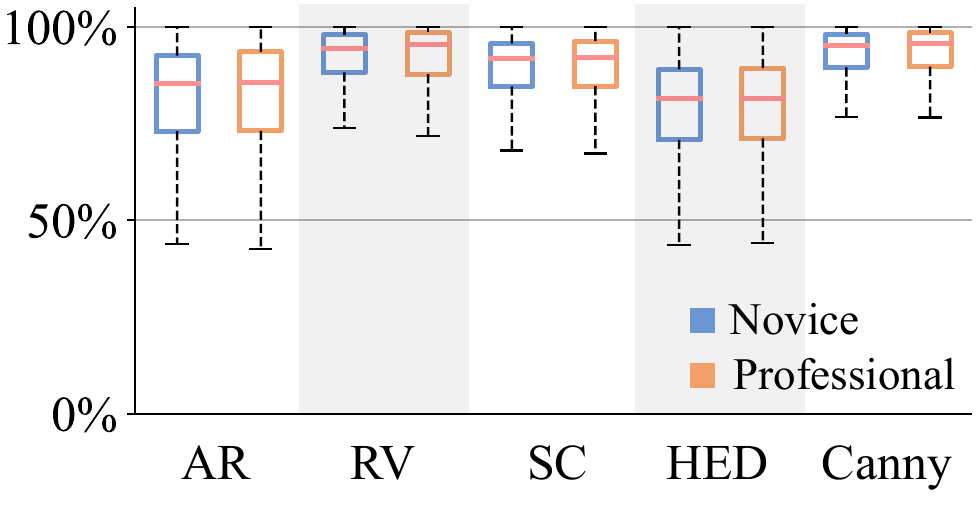}
        \end{minipage}
    }
    \caption{Comparison between the algorithm-generated line drawings and the freehand sketches with different registration strategies. (a) shows qualitative comparisons of two sets of drawings by different methods. We dilated the generated results for clear demonstration in this figure. For the user sketches shown on the right, the three columns (from left to right) represent the sketch-level, stroke level, and pixel-level registered versions of the original sketches, respectively; the upper row and the lower row illustrate sketches from novices and professionals, respectively. (b), (c), and (d) illustrate the precision (Top) and recall (Bottom) comparisons between different levels (sketch-level, stroke-level, and pixel-level, respectively) of the registered freehand sketches and the algorithm-generated drawings.
    }
    \label{fig:CGDA_quan}
    }
\end{figure*}

\subsection{How Similar are Freehand Drawings and Synthetic Sketches?}
\label{sec:cgda}

To facilitate potential applications of designing more effective algorithms for synthesizing freehand-like sketches, we 
analyzed the similarity between our collected freehand sketches (professional vs. novice) and line drawings generated by representative line drawing generation algorithms.
We computed the precision $P$ and recall $R$ defined in Eq. \ref{eq:PR} for the generated drawings of one human drawing under different registration strategies.
We rasterized the vector sketches as one pixel width in the three levels of registration schemes as three groups of targets and considered the synthetic sketches to capture them according to the user drawing skills (professional vs. novice).
We chose five representative algorithms for line drawing generation from two categories: 
NPR-based methods (apparent ridges (AR) \cite{judd2007apparent}, ridges and valleys (RV) \cite{ohtake2004ridge}, and suggestive contours (SC) \cite{decarlo2003suggestive}) and edge detection-based methods (Canny \cite{canny1986computational} and holistically-nested edges (HED) \cite{xie2015holistically}).
We choose thresholds of the algorithms such that the number of generated pixel was the closest to the median number of filled pixels in user drawings, as done in \cite{Wang2021Tracing}.
\ac{Figure \ref{fig:CGDA_quan} (a) shows the produced sketches and our registered drawings. We present the precision and recall on all the drawings using box plots to show the stability and plausibility of the algorithms on diverse models in Figure \ref{fig:CGDA_quan} (b), (c), and (d) for sketch-level, stroke-level, and pixel-level performance, respectively.}

\begin{figure*}[t]{
    \centering
    \subfloat[]{
      \includegraphics[width=.7\linewidth]{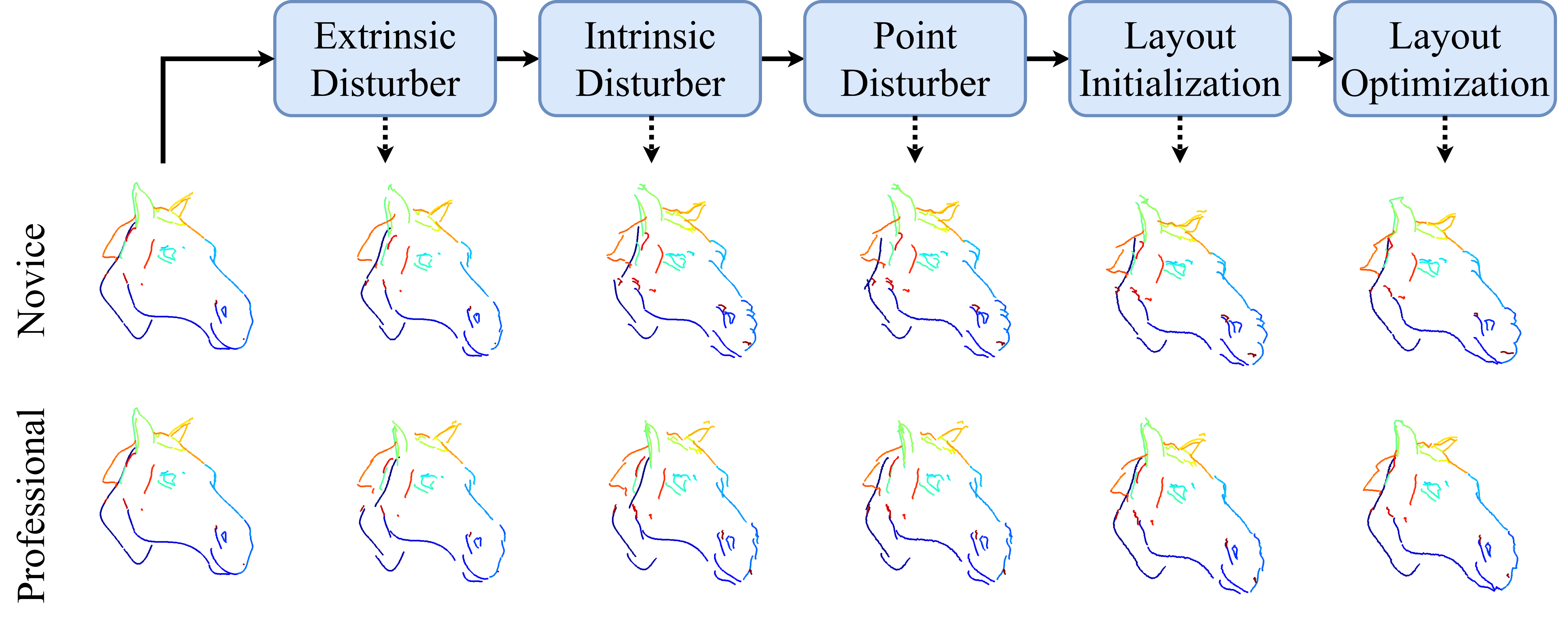}
    }
    \subfloat[]{
      \includegraphics[width=.29\linewidth]{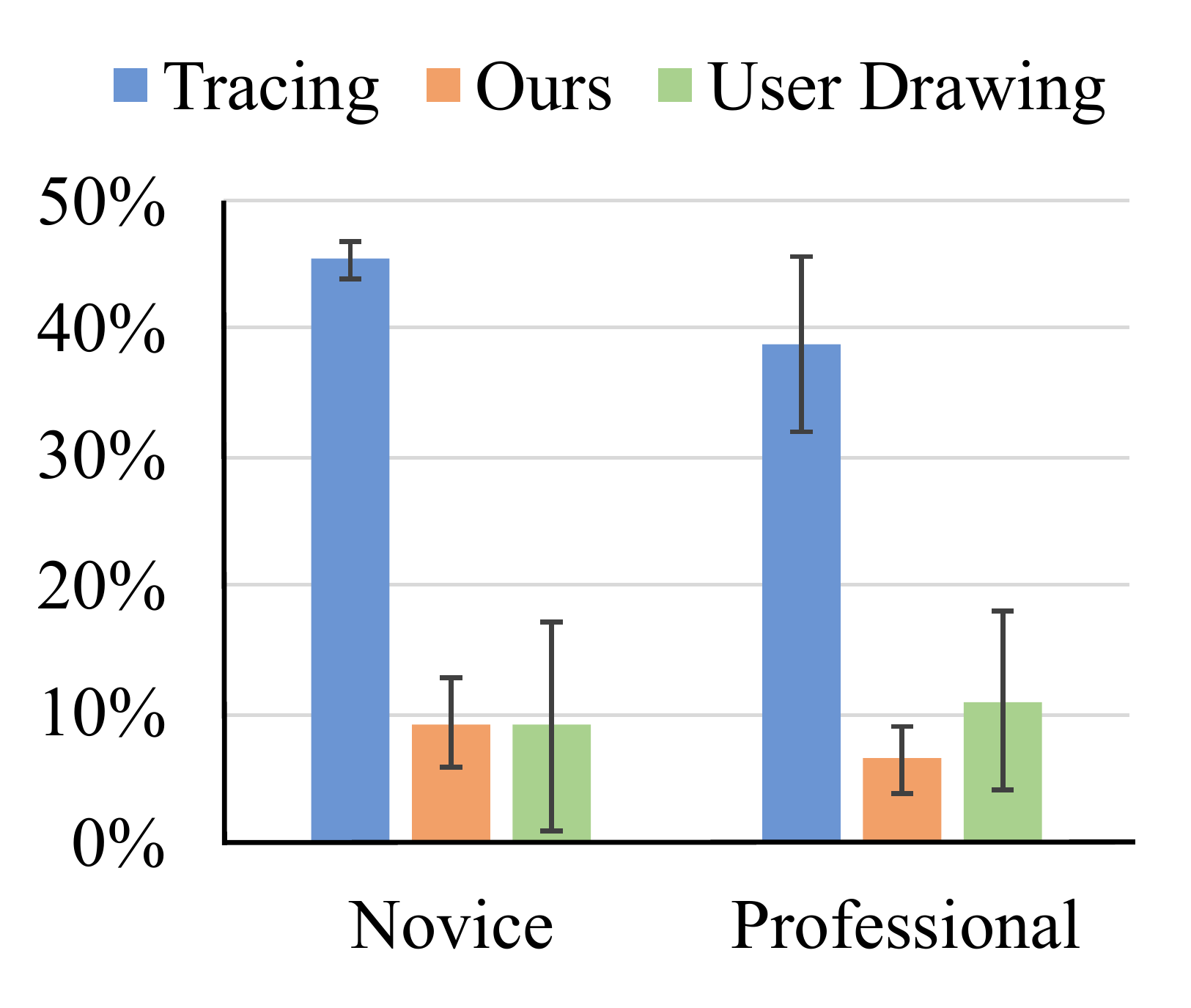}
    }
    \caption{(a) The pipeline of our freehand-style sketch synthesis method. The two rows of images show novice-style and professional-style generation, both of which are fed with the same noise levels ($n_1$=$n_2$=0.3). Each stroke of the sketches are color-coded to highlight their changes. Please zoom-in to find out the details at each step. (b) The average voting rate of the perception study \ac{({where} the participants selected the least-plausible human-drawn sketches)} for the three types of sketches.}
    \label{fig:syn_pipeline}
    }
\end{figure*}
\textbf{Conclusion}:  
The professional sketches are more easily approximated by the synthetic drawings, but for the novice sketches, more efforts are still needed. \ac{The NPR-based methods sometimes lose the boundary lines or produce excessive lines, depending on the quality of 3D meshes. The Canny algorithm was the most stable to match the freehand drawings and synthetic ones.}
Considering the expressiveness and feasibility, the stroke-level registered sketches could be good targets for generating freehand sketch-like drawings using \ac{data-driven} sketch synthesis algorithms.

\section{Applications}

\ac{Our main goal is to understand how differently novices and professionals sketch 3D objects and quantify their key differences. The analysis (Section \ref{sec:analysis}) thus provides quantifiable proofs to benefit various applications (e.g., sketch synthesis and future research on sketch-based modeling), instead of directly improving sketch-based 3D modeling for novices.}
In this section, we first present a new approach for freehand-style sketch synthesis (Section \ref{sec:sketch_synthesis}) and then show a preliminary evaluation of existing sketch-based 3D reconstruction on our dataset as a testing set (Section \ref{sec_3d_reconstruction}).

\subsection{Freehand-style Sketch Synthesis}
\label{sec:sketch_synthesis}

Although tracings have some similar features to freehand sketches \cite{Wang2021Tracing}, they could not substitute user drawings completely due to the difference from global to local (Section \ref{sec:factors}), hindering applying tracings or tracing-like sketches (e.g., contours of objects) to 
sketch-based applications expecting freehand sketches as input. To narrow the gap between them, we propose a learning-based framework to produce freehand-style sketches from tracings based on the collected drawings and their multi-level registered results. As shown in Figure \ref{fig:syn_pipeline} (a), we first reshape each stroke from global to local, i.e., disturbing its 
extrinsic and intrinsic parameters by three disturbers, and then re-group strokes following the relationship across them by layout refinement. The details of our synthesis method are as follows:

\textit{Preprocessing.} We first fit the tracing strokes with unequal length to B\'ezier curves with fixed six control points as inputs. 

\textit{Stroke Extrinsic Disturbing.} We adopt an extrinsic disturber, i.e., a multilayer perceptron (MLP) with two hidden layers in our implementation, to predict a similarity transformation ($R$, $T$, $S$) for each stroke. The disturber is trained on the paired data, with each pair containing a sketch-level registered sketch and its corresponding stroke-level registered sketch, taking a B\'ezier curve as input and the transformation parameters $R_L$, $T_L$, $S_L$ (Section \ref{sec:factors}) as ground truth.

\textit{Stroke Intrinsic Disturbing.} Next, the extrinsically reshaped B\'ezier curves are fed into an intrinsic disturber to approximate freehand-style stroke paths. The disturber (an MLP similar to that used for the extrinsic disturber) is trained on the paired pixel-level (input) and stroke-level (ground truth) results, with each stroke fitted to a B\'ezier curve.

\textit{Point Disturbing}. Since the B\'ezier curve is too smooth to be like a human-drawn stroke with flutter points, \ac{we adopt Gaussian noise to model the random jittery and} train one more disturber (an MLP) to add a normal noise $N(\mu,\sigma^2)$ into the rendered B\'ezier curve. 
The training dataset is paired sketch-level registered strokes and their fitted B\'ezier curves.

\textit{Layout Initialization.} After the  extrinsic and intrinsic disturbing, the changed strokes are messy spatially on canvas and thus hard to \ac{perceive} a reasonable human-drawn drawings, as shown in Column 4 of Figure \ref{fig:syn_pipeline} (a). To refine the layout, we first translate the strokes one-by-one (similar to a stroke-by-stroke drawing process of a human artist) according to the positions of the previous strokes in the tracing layout, following
\begin{equation}
    d_i=d_i+\sum_{j<i}^jw_j\cdot(d_i^\beta-(t_i^\beta-t_j^\alpha+d_j^\alpha)),
\end{equation}
\begin{equation}
    w_j=softmax_{j<i}(1/(dist(t_i^\alpha,t_j^\beta)+1)),
\end{equation}
where $t_i$ and $d_i$ respectively indicate the $i$-th tracing stroke and the disturbed stroke, $\alpha$ and $\beta$ are the point indices of the two closest points between the two strokes $t_i$ and $t_j$, and $dist(\cdot)$ is the Euclidean distance between the two points. 

\textit{Layout Optimization.} With layout initialization for each stroke, the produced sketches have a stroke layout similar to the input tracings, but some strokes connected with each other in the tracings are still broken (Column 5 of Figure \ref{fig:syn_pipeline} (a)). To address the issue, we perform curve optimization on the problematic strokes, as done in \cite{ye20213d}. The objective function is defined as
\begin{equation}
    F(d_i)=T_p+w_sT_s+w_mT_m,
\end{equation}
where $T_p$ is the position term to guarantee the two strokes connected in the tracing stay connected, while the shape term $T_s$ and the smooth term $T_m$ help preserve the disturbed shape and stroke smoothness, respectively. Please refer to the paper \cite{ye20213d} for a more detailed definition for each term. The optimized results (Figure \ref{fig:syn_pipeline} (a), Column 6) show that the optimization can effectively fix the broken issue.

We also introduce an extra input variable $n_1$
for the extrinsic disturber to control the noise level by computing the change magnitude between input and output. A similar variable $n_2$ is introduced for the intrinsic disturber. 
To simulate novice-style and professional-style drawings, we had the disturbers (i.e., the MLPs) respectively trained on the novice and professional datasets. 
Figure \ref{fig:syn_pipeline} (a) shows examples of the synthesized results.

We conducted a perception user study to evaluate whether our synthesized results are close to human-drawn sketches or not. 
The evaluation was done by inviting participants to complete an online questionnaire.  We prepared two sets of questions respectively for novices and professionals. Each set contained 9 groups, corresponding to the 9 categories we used for data collection. Each group corresponded to a prompt randomly picked from the concerned category, and consisted of its corresponding tracing, five user drawings (from either novices or professionals), and one synthesized result.
\ac{Note that before the study, we had conducted a similar but preliminary study showing only one user drawing (randomly picked), but found people had 
different senses of recognizing freehand drawings, i.e., some preferred ``noised-up” drawings while others preferred ``aligned-well” ones. 
Instead of informing what the computer-generated sketches are like, for a fair comparison, in the final study we showed all the five drawings from the same group for users to form consistent criteria by themselves.}
We sampled the noise level $n_1$ and $n_2$ within [0,0.2] to synthesize freehand-style sketches. We asked each participant to pick out two options that were the least possible to be human-drawn sketches from the seven items given the corresponding image prompt. 
48 participants helped with the study. Most of the participants had no professional drawing experience. 
In total, we got 48 (participants) $\times$ 18 (questions) = 864 subjective evaluations. 

\begin{figure}[htb]{
    \centering
    \includegraphics[width=\linewidth]{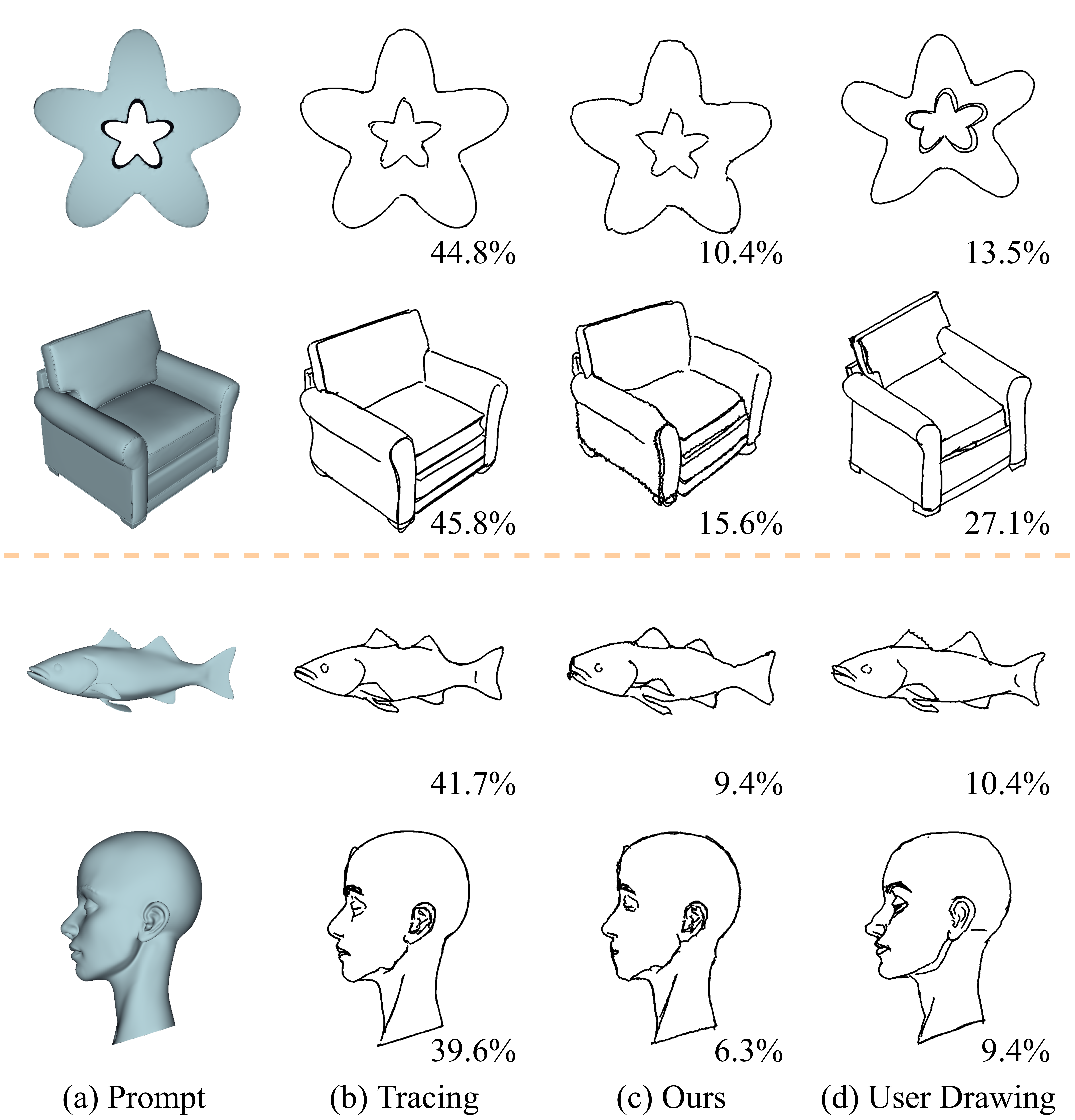}
    \caption{Examples of the perception study. (d) is the highest voted user drawings among the five user drawings in each group in the study. The top two rows are for the novice-style comparison, while the bottom two rows are for the professional-style comparison. The percentage value
    at the bottom-right corner of each drawing is the averaged voting rate \ac{(of the participants for selecting the least-plausible human-drawn sketches)} over 48 participants for a certain type of sketches.}
    \label{fig:syn_example}
    }
\end{figure}
Figure \ref{fig:syn_pipeline} (b) visualizes the average voting rate for the three sketch types of options over 9 object groups for the novice and professional groups separately. 
Note that each human-drawn sketch got a voting rate and we computed the average voting rate of human-drawing type among the five drawings for each group.
It is obvious that for both the novice and professional types, the tracings were easily recognized by the users, while the generated results had similar voting rates with the human drawings, meaning that our method could significantly fool the participants. The U-tests on all the 18 groups also support our results have a significant difference from tracings ($p$=1.4e-7) but not from user drawings ($p$=0.32). 
Figure \ref{fig:syn_example} shows some examples of the perception study. Please find more implementation details and generated examples in \ac{Section 4.1 of Supplement}. 

\begin{table*}
\caption{Quantitative evaluation of the results by four single-view reconstruction methods taking either the novice drawings (indicated by ``N'' after each method name) or the professional drawings (indicated by ``P'') as input. ``IoU'' and ``Normal'' are not applicable for PSGN results.
}
\label{tab_diff_mtds}
\begin{tabular}{c|c|c|c|c|c|c|c|c}
\hline
Method & Pixel2Mesh (N) & Pixel2Mesh (P) & R2N2 (N) & R2N2 (P) & PSGN (N) & PSGN (P) & Occ-Net (N) & Occ-Net (P) \\
\hline
IoU $\uparrow$ & 0.146 & 0.154 & 0.242 & 0.260 & - & - & 0.280 & 0.289\\
CD ($\times 10^{-3}$) $\downarrow$ & 4.342 & 4.047 & 3.135 & 3.493 & 2.168 & 1.911 & 2.613 & 2.846 \\
NC $\uparrow$ & 0.620 & 0.628 & 0.584 & 0.598 & - & - & 0.728 & 0.734 \\
\hline
\end{tabular}
\end{table*}
\subsection{Preliminary Evaluation of Sketch-based 3D Reconstruction} 
\label{sec_3d_reconstruction}

\ac{While it seems obvious that existing sketch-based 3D reconstruction methods would perform better with professional sketches, the reconstruction performance gap between professional and novice sketches has not been seriously studied before.}
Since we collected a dataset of freehand sketches from the two groups for 3D objects, one direct usage of our data is to test the performance of existing sketch-based 3D reconstruction methods.
Although the volume of data we collected is moderately large (compared to the existing stroke-level sketch datasets), the samples in our dataset are far from sufficient to train any deep learning-based models from scratch.
We thus performed a demonstrative experiment to show the simplest way of utilizing our collected data as testing samples in the shape reconstruction task. 
For simplicity, we focused on a single-view reconstruction (SVR) task tested on the \emph{Chair} category, since \emph{Chair} is known for its large variations in both geometry and structure. 

\ac{We chose four representative image-based shape reconstruction methods taking single-view sketches as input, including Pixel2Mesh \cite{wang2018pixel2mesh}, 3D-R2N2 \cite{choy20163d}, Occ-Net \cite{mescheder2019occupancy}, and PSGN \cite{fan2017point}. Each of them respectively sets meshes, voxels, implicit functions, and point clouds as a 3D shape representation. For a fair comparison, we trained the four models with synthetic sketches from \cite{zhong2020deep} and then tested them {on 300 pairs of novice and professional sketches} to obtain 600 reconstructed models 
for each method.}

\ac{To compare the reconstruction performance of the four methods, we adopted three evaluation metrics to quantitatively measure the difference between the reconstructed meshes and the ground truth, including 3D Intersection over Union (IoU), Chamfer distance (CD), and Normal Consistency (NC) scores, as used by Mescheder et al. \shortcite{mescheder2019occupancy}. Table \ref{tab_diff_mtds} presents the results averaged over the sketches from the novices and the professionals, separately. It shows that the reconstruction results from the professional sketches generally obtained better metric values. This is possibly because professional sketches are more similar to the synthetic sketches used in the training stage. 
PSGN received the lowest CD among all the four methods, mainly because CD was the only loss for used for training {PSGN}. Exceptionally, R2N2 and Occ-Net performed better on the novice sketches in terms of CD. We attribute this to no Chamfer loss being employed when training the two methods. Figure \ref{fig_chairs} shows qualitative results consistent to the metric values. Please find more setting details and visual comparisons in Section 4.2 of Supplement.}

\begin{wrapfigure}{r}{.5\linewidth}
\vspace{-13pt}
\hspace{-18pt}
\includegraphics[width=1.12\linewidth]{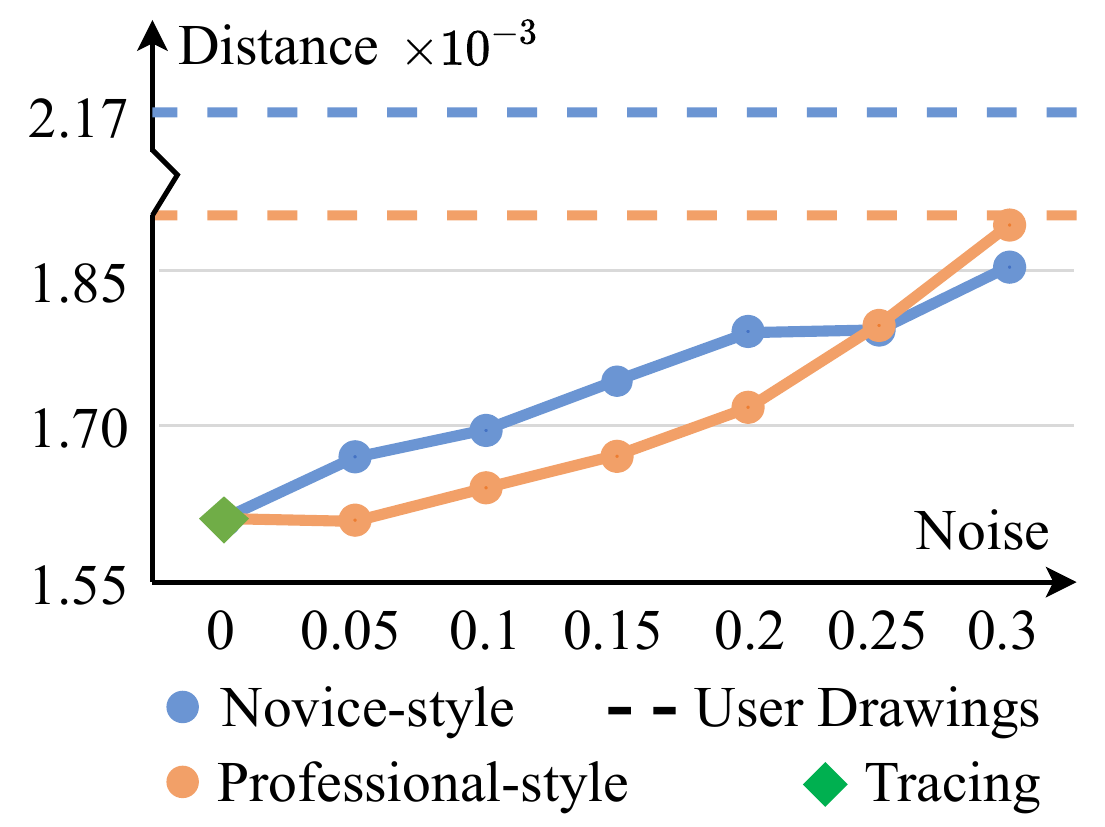}
\vspace{-12pt}
\end{wrapfigure}
To further evaluate the effect of noise level on the synthesized sketches (Section \ref{sec:sketch_synthesis}),
we tested the PSGN model with the synthesized sketches 
under different noise levels considering its consistent performance and simple input requirement (R2N2 and Occ-Net present inconsistent evaluation results while Pixel2Mesh requires an additional view angle input). 
We prepared the synthetic sketches with the noise levels $n_1$=$n_2$ ranging from 0.05 to 0.3 with step size of 0.05 and calculated the CD values of the reconstructed results \ac{given the sketches of the two-group styles.} 
Since sketches with noise larger than 0.3 exhibit severe stroke deformation, failing to preserve the target shape information, we exclude such sketches in this evaluation.
We also tested the sketches with 0 noise level (i.e., tracings) using the same scheme. 
See the inset figure for the results. 
The two groups of synthetic sketches show an increasing trend with the rise of noise level.
Generally, the synthetic professional sketches are more like the sketches used in training, showing lower CD values before noise level 0.25.
All the synthesized sketches present lower CD values than our collected freehand sketches i.e., the dashed lines in the inset figure (professional $1.91 \times 10^{-3}$ and novice $2.17 \times 10^{-3}$), indicating that freehand sketches are more challenging for reconstruction techniques and needed for understanding their actual performance. 

\begin{figure}[t]
    \centering
    \includegraphics[width=\linewidth]{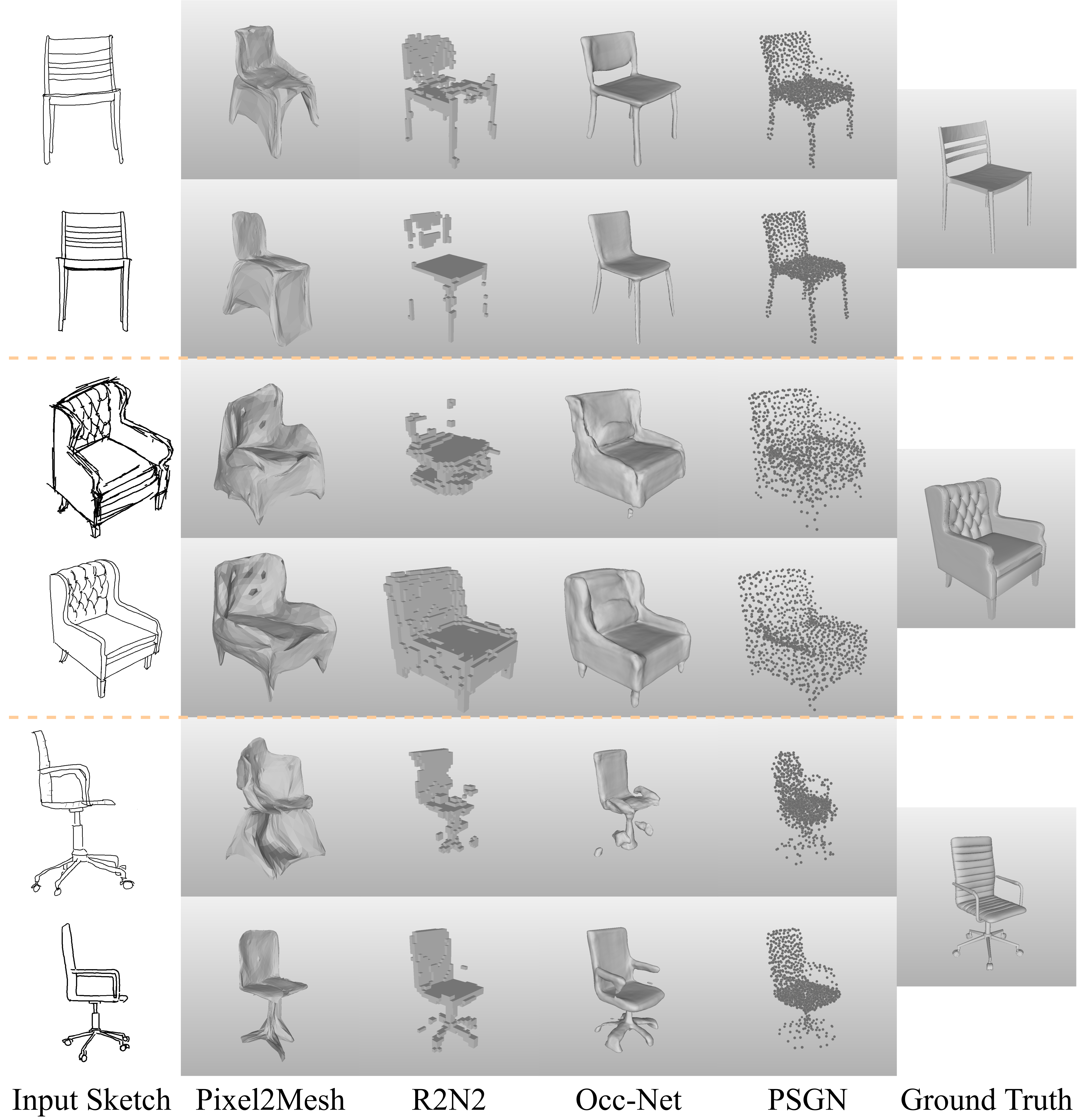}
    \caption{3D reconstruction results given sketches by the novice users (Top of each group) and the professional users (Bottom of each group). 
    The ground truth models are illustrated on the right of each group. Results are best viewed with zoom-in.
    }
    \label{fig_chairs}
\end{figure}

\section{Conclusion and Discussion}

In this paper, we conducted a study to investigate how differently professional and novice users sketch 3D objects on a newly collected dataset, called \emph{\datasetName}.
By three levels of analysis, we found that
(a) the professionals drew 
sketches more precisely as to the global variance (translation, rotation, scale) than the novices;
(b) the professionals depicted strokes more fit to the underlying shapes than the novices; 
and (c) both the professional and novice users tended 
to depict the same contents given the same prompts, yet the professionals drew more consistently with each other and the novices' depictions of the contents were 
more varied.
By comparing the freehand sketches to various algorithm-generated drawings, we found converting the synthetic drawings to freehand-like ones by the guidance of stroke-level registered sketches are more effective and feasible. \ac{We believe the quantitative analysis could enlighten a new norm for sketch synthesis measurement.}

We proposed a novel framework for freehand-style sketch synthesis trained on our collected dataset and the multi-level registered results. The evaluation shows our synthesis method and dataset are promising to be applied to 
(a) sketch style transfer between novices and professionals
\ac{given the conclusion of Section \ref{sec:common}};
(b) model- or image-to-freehand sketch tasks by incorporating our method with the synthesis algorithms (e.g., those in Section \ref{sec:cgda}); 
(c) sketch data augmentation, i.e., creating a large-scale sketch dataset to boost the performance and particularly generalization ability of sketch understanding techniques, e.g., for sketch-based shape reconstruction and sketch segmentation; etc. 
\ac{Also, it is interesting and potential to utilize deep learning techniques with our freehand-style sketch synthesis framework to produce higher-quality freehand drawings for diverse objects and introduce more types of disturbers (e.g., viewpoint) to obtain varied results.}

We used our collected data as a testing set for preliminary evaluation of several state-of-the-art 3D reconstruction methods, and found that the professional sketches produced results with higher quality than the novice sketches \ac{by quantitative metrics}.
We directly applied our collected sketches to the generation modules to get the reconstruction results, which may suffer from the domain gap between the sketches we collected and the data that the generation models were trained on.
In the future, we are interested in using our sketch synthesis method to create a large dataset of freehand-style sketches for re-training or fine-tuning the pre-trained networks. This approach could potentially improve the performance of current sketch-based 3D reconstruction methods.

We implemented the data collection user interface on tablets with styluses. 
It is convenient to record and process the data digitally while this setting may potentially incorporate the inaccuracies for non-digital drawing tool users to reflect their real drawing skills.
Specifically, one artist claimed that she could not perform as well on the tablet as in the paper-and-pencil setting.
In the sketch collection stage, we required the participants to depict the shape in the prompt images and ignore the lighting/shading effects.
This limits our analysis only on the perception and depiction of shape information from the collected data.
The analysis on perception and depiction of lighting/shading effects for people is an interesting subject and worthy of further efforts towards this direction, like \cite{Wang2021Tracing}. 

\ac{For collecting relatively large-scale data, we chose the algorithmic registration to save enormous labor of the manual alignment as done by Cole et al. \shortcite{cole2008people}. Although we believe our method (Section \ref{sec_registration}) could preserve most of users’ intentions in a low-cost way, it is potentially the best way to ask users to confirm whether their intentions are faithfully maintained after registration.}
The registration algorithm works smoothly for most of the sketches in our dataset. For the badly registered samples, we recruited users to manually assign the corresponding points of the sketches and the prompt images to provide initialization in the registration process \ac{(see Section 2 in {Supplement}}. 
Despite all the efforts we made, there are still some samples that cannot be registered precisely (empirically, $E^*$<1.2 in Eq. \ref{eq:E}) to the prompts. Since the registration errors lead to the inaccuracies in our quantitative statistics, we excluded such samples in our dataset. There are some sketch samples with strokes that mismatch with the prompts severely in our dataset. Such sketches only take less than $4.6\%$ in our dataset. We hope to use more sophisticated registration methods to resolve such a problem. 

In the future, our dataset can be further processed or annotated to serve as new benchmarks for various sketch understanding tasks. 
With the rich correspondence between our sketches and the 3D models, the inter-sketched tasks requiring 3D information previously could be evaluated effectively, such as sketch segmentation \cite{yang2021sketchgnn,huang2014data}, multi-view sketch correspondence \cite{yu2020sketchdesc}.
By analyzing the difference between the drawings from novices and professionals, our study could be insightful for art training, as well as applications aiming at assisting the drawing process, like \cite{dixon2010icandraw}.
\ac{Our dataset can potentially help design more effective sketch-based modeling techniques for novice users, e.g., by utilizing more characteristics (like {the} stroke order) of sketches.}

Besides, we summarize some future directions that could be possibly done with our current study. 
We did not set the time limit for completing one drawing in our current study. It is interesting to study the drawings with given a time limit and compare the drawing differences from people with different  drawing skills.
It is also interesting to study the perspective perception and depiction of users with different levels of drawing skills, with potential annotations on the 3D models as fiducials.
\ac{In addition, we only adopted some low-level metrics to evaluate the accuracy of strokes and sketches in the analysis. It would be insightful to devise some high-level metrics, e.g., line parallelism, roundness, curvature monotonicity, to further study the difference of the drawings.}

\begin{acks}
We thank the anonymous reviewers from SIGGRAPH 2022 and SIGGRAPH Asia 2022 for the constructive comments, and the sketch data contributors for their time and effort.  
A special thank goes to Aaron Hertzmann, who provided great support to this project in its early stage. 
The work was supported by an unrestricted gift from Adobe, and grants from the Research Grants Council of the Hong Kong Special Administrative Region, China (No. CityU 11212119), City University of Hong Kong (No. 9229094), Beijing Nova Program of Science and Technology (No. Z191100001119077), and the Centre for Applied Computing and Interactive Media (ACIM) of School of Creative Media, CityU.
\end{acks}

\bibliographystyle{ACM-Reference-Format}
\bibliography{new_bib}



\section*{Supplementary material}

# Supplemental Material
# DifferSketching: How Differently Do People Sketch 3D Objects?

CHUFENG XIAO*, School of Creative Media, City University of Hong Kong, China
WANCHAO SU*, School of Creative Media & Department of Computer Science, City University of Hong Kong, China
JING LIAO, Department of Computer Science, City University of Hong Kong, China
ZHOUHUI LIAN, Wangxuan Institute of Computer Technology, Peking University, China
YI-ZHE SONG, SketchX, CVSSP, University of Surrey, UK
HONGBO FU†, School of Creative Media, City University of Hong Kong, China

## 1 DATASET OVERVIEW

Please find the thumbnails of all the collected sketches as well as the corresponding prompts and the multi-level reigister results for the 9 categories in the accompanying file folder. For a thumbnail of each category, except for the first column (rendered image prompts and tracings), the left groups of five sketches are from the novices, while the right groups of five drawings are from the professionals, including the drawings and their corresponding registered results on sketch-level, stroke-level, pixel-level (from top to bottom) on the right side.

## 2 MULTI-LEVEL REGISTRATION

We implemented our pixel-level registration based on the point-to-point registration using SimpleITK [Yaniv et al. 2018] following some settings of Wang et al. [Wang et al. 2021]. In the initialization stage for a displacement field, we applied global affine transformations and B-Spline transformations at four cascaded scales of the images. B-Spline transformation is a deformable transform over a bounded spatial domain using a B-Spline representation for 2D coordinate space, of which the parameters are default values in SimpleITKv4 [Yaniv et al. 2018]. We applied the automatic-initialized displacement field to the sketches one by one and filtered out the failure cases, each of which is far away from the target center or with messy lines. Similar to Wang et al. [2021], we recruited another group of four users to manually assign the corresponding points between the freehand sketches and the corresponding prompt images for the sketches where the automatic initialization failed. We set 4-39 corresponding points for each of 745 freehand sketches and fitted them to thin plate spline models for the semi-automated initialization of the badly initialized sketches. At the optimization stage (`ImageRegistrationMethod()` in SimpleITK), we performed the optimization to get the optimal displacement maps by maximizing the correlation between the sketches and the tracings using stochastic gradient descent optimizer.

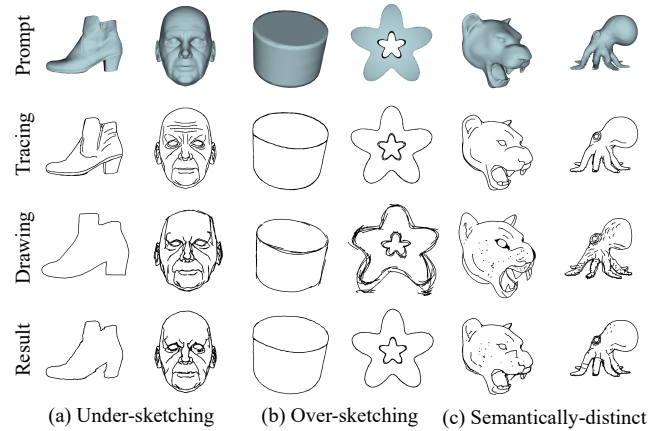

Fig. 1. Examples of our proposed pixel-level registration for the three types of special cases. Please find the corresponding progressive results in Figure 2. The sketches (from left to right) were drawn respectively by *Novice 49, Professional 38, Novice 3, Professional 6, Professional 18, and Professional 34.*

We implemented pixel-level registration by incorporating the point-to-point registration and iterative rasterize-and-optimize process, As shown in Figure 1, there exist three special cases in our collected drawings and targeted tracings: a) Under-sketching: a drawing could not cover all of the shape information as the corresponding tracing does; b) Over-sketching: a drawing repeats several strokes to depict the same shape region; c) Semantically-distinct: a drawing contains strokes without semantic correspondence with any line of the corresponding tracing. The "Result" row of Figure 1 shows that our proposed pixel-level registration method performs well for these special cases. Please find the registration progress in Figure 2.

## 3 DATA ANALYSIS

### 3.1 Basic Statistics

We conducted the statistical analysis on the collected freehand sketches in both spatial and temporal perspectives, within and across the two user groups. We first report the basic features of our analysis. The average numbers of strokes for novices and professionals in our dataset were 70.45 and 105.00, respectively. The average time spent in drawing a sketch was 3.40 minutes for novices and 4.48 minutes for professionals. Professionals used the undo function more often

*Authors contributed equally.
†Corresponding author.

Authors' addresses: Chufeng Xiao, School of Creative Media, City University of Hong Kong, China, chufeng.xiao@my.cityu.edu.hk; Wanchao Su, School of Creative Media & Department of Computer Science, City University of Hong Kong, China, wanchao.su@cityu.edu.hk; Jing Liao, Department of Computer Science, City University of Hong Kong, China, jingliao@cityu.edu.hk; Zhouhui Lian, Wangxuan Institute of Computer Technology, Peking University, China, lianzhouhui@pku.edu.cn; Yi-Zhe Song, SketchX, CVSSP, University of Surrey, UK, y.song@surrey.ac.uk; Hongbo Fu, School of Creative Media, City University of Hong Kong, China, hongbofu@cityu.edu.hk.

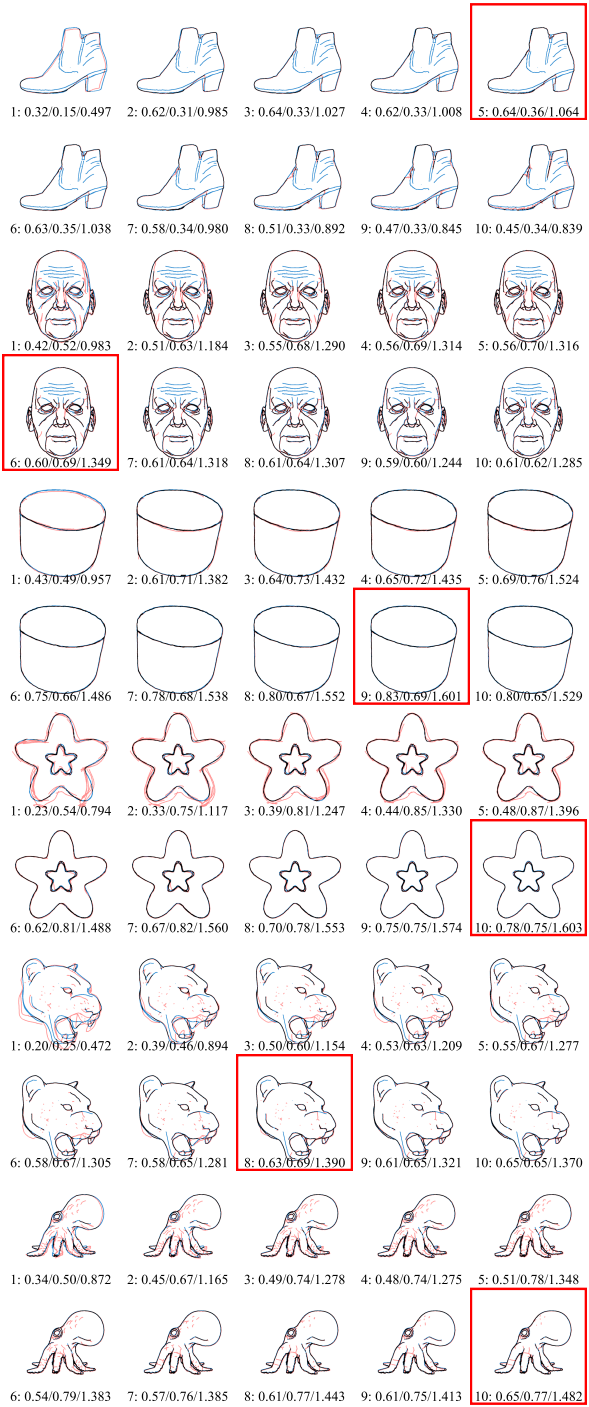

Fig. 2. The progressive results by our proposed pixel-level registration approach for Figure 1. Each image visualizes the registered results of each iteration, where the drawing pixels that have not been registered correctly are in red and the tracing pixels to be registered by the drawings are in blue, while the pixels overlapped by them are in black. The bottom text of each image indicates the performance evaluation ("$i$-th iteration: $E_i$ / $P_i$ / $R_i$"), by which we picked the optimal iteration as the final result (red bounding box).

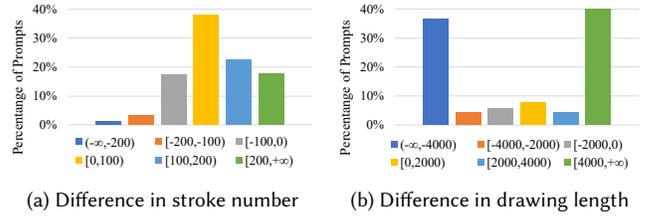

(a) Difference in stroke number  (b) Difference in drawing length

Fig. 3. Two histograms of the differences between the drawings by professional and novice users across all the categories on stroke number and drawing length (accumulated stroke length). The differences are computed by subtracting the total value of five novice users from that of five professional users for the same prompt (P-N).

than novices (17.75 times versus 8.79 times on average). By calculating the pause duration proportions for each drawing, we found that during sketching, all the users spent more time pausing than drawing. Specifically, novice users spent 64%-82% of the time pausing when creating sketches for different categories of objects; and professional users's pausing time ranged from 70% to 83%. Professional users spent slightly more time on pausing than novice users on average (76% vs. 70%). All the users drew with similar strength, with a slightly milder average pressure of 0.40 for the professional drawings (novice as 0.47, range: 0 (no pressure) - 1 (hardest)). Please find more categorical statistics in Table 1.

By comparing the drawings in terms of groups using the category-level statistics, we found that the professional users drew more strokes for the prompts in most categories (7 out of 9). For categories with sharp shapes, i.e., *Chair*, *Industrial Component*, *Lamp*, *Primitive*, the novice users drew longer paths for the majority of the prompts. For the prompts with complex geometries, the professionals drew paths longer than the novices. We computed the difference of accumulated numbers of strokes between the professional and the novice groups for all the drawings. The results are shown in Figure 3 (a). Generally, the professionals drew more strokes for over 78% of the prompts. As to the accumulated stroke length, it shows a bimodal trend (see Figure 3 (b)): for 41% of the prompts, the novices drew more pixels (>4000 pixels) than the professionals, yet for other 41% prompts, the professional users drew more pixels (>4000 pixels). Compared to the novice group, professional users preferred to draw longer paths for complicated models, e.g., *Animal*, *Human Face*, *Vehicle*, etc.

**Conclusion**: The professional users tended to draw more strokes and spend longer time than the novice users given the same prompts and the professionals modified the drawings more frequently than the novices, especially for the models with richer details.

### 3.2 Where Does Key Difference Occur When Sketching?
From a sketch-level perspective, over 55% of the drawings from the two groups require small rotation ($R_G \leq 4°$) to match the prompts globally. It shows that people tended to sketch 3D objects globally similar to the prompt orientation, and the tendency is stronger in the professional group. There are more professional drawings with the translation distance under 100 pixels (69% drawings) and the

Table 1. Extended categorical statics for our dataset. We report the comparisons between the drawings by professional (P) and novice (N) users in terms of the number of strokes and the path length. We compute the ratio of the prompts in each category whose corresponding professional sketches contain more strokes than the novice sketches. Similarly, we also report the ratios for stroke path length (in pixels), the average time used for drawing sketches within each category, the ratio of pause duration over the whole drawing time, and the average times of using the undo function. For the last three statistics, we report the values in $novice/professional$ pairs.

| Category | Animal | Animal Head | Chair | Human Face | Industrial Cpnt | Lamp | Primitive | Shoe | Vehicle |
|---|---|---|---|---|---|---|---|---|---|
| Stroke Number (P>N) | 0.30 | 0.75 | 0.75 | 0.61 | 0.8 | 0.76 | 0.30 | 0.78 | 1.00 |
| Path Length (P>N) | 0.77 | 0.53 | 0.45 | 0.67 | 0.06 | 0 | 0.45 | 0.73 | 0.98 |
| Time (minutes) | 3.10/4.91 | 3.09/3.49 | 3.45/4.37 | 8.25/11.02 | 3.06/3.82 | 3.59/2.83 | 1.97/2.16 | 2.97/4.23 | 3.32/6.18 |
| Pause Proportion | 0.69/0.75 | 0.74/0.74 | 0.68/0.78 | 0.82/0.83 | 0.69/0.70 | 0.70/0.72 | 0.75/0.79 | 0.64/0.76 | 0.71/0.77 |
| Undo Count | 7.38/20.59 | 6.77/12.28 | 11.46/24.87 | 22.67/42.04 | 7.61/21.96 | 7.23/10.58 | 14.48/12.87 | 5.51/12.50 | 7.28/15.24 |
| Pressure Average | 0.46/0.44 | 0.49/0.41 | 0.45/0.43 | 0.46/0.43 | 0.54/0.41 | 0.49/0.41 | 0.44/0.30 | 0.46/0.36 | 0.45/0.43 |
| Pressure Std | 0.10/0.10 | 0.11/0.10 | 0.10/0.10 | 0.11/0.11 | 0.11/0.09 | 0.11/0.10 | 0.09/0.08 | 0.10/0.09 | 0.09/0.10 |

Table 2. The statistics for three levels of registration differences to the corresponding tracings in terms of Rotation ($R$), Translation ($T$) and Scale ($S$). We show $mean/standard\ deviation$ pairs on the two groups for each metric in three analysis levels. Note that we only evaluated the valid drawings / strokes that are registered correctly. The valid counts are also presented here. We have '-' for the pixel-level rotation and scale, since these values cannot be evaluated at the pixel level.

| Level | Skill | Count | $R$ (°) | $T$ (pixel) | $S$ |
|---|---|---|---|---|---|
| Sketch-level | N | 1680 | 5.06 / 5.52 | 121.03 / 89.76 | 0.89 / 0.20 |
|  | P | 1778 | 3.57 / 4.58 | 86.68 / 70.11 | 0.95 / 0.14 |
| Stroke-level | N | ∼59.9K | 12.83 / 14.54 | 222.58 / 198.23 | 1.02 / 0.39 |
|  | P | ∼89.4K | 10.60 / 11.98 | 184.53 / 172.22 | 1.01 / 0.32 |
| Pixel-level | N | ∼11.2M | - | 8.67 / 14.14 | - |
|  | P | ∼12.6M | - | 5.17 / 8.94 | - |

scaling factor falling into [0.9, 1.1] (75% drawings), compared to the novice drawings (48% and 21%, respectively). The difference of scaling factor indicates that the novice users had diverse scaling settings (a more gentle distribution) for drawing, compared to the professionals who tended to have their drawings with similar scales to the prompt targets.

From a stroke-level perspective, in the professional group, there are 64% strokes with $R_L \leq 10°$, 35% strokes with the translation distance error with $T_L \leq 100$, and 45% strokes with the scaling error in $S_L \in [0.9, 1.1]$. In the novice group, the numbers of strokes falling into the correct ranges of the three metrics are all less than those in the professional group, i.e., 57% for $R_L$, 26% for $T_L$, and 23% for $S_L$. It reflects that the professional users were more capable of placing strokes with the proper position, orientation, and especially scale than the novice users. Thus, it is challenging for novice users to make part-by-part ratios stay consistent with each other.

For the pixel level, it is obvious that the professional users drew more pixels (63%) near the correct position (≤4 pixels) and fewer outlier pixels (17%) (≥8 pixels), compared to the novice users who drew 49% and 29% pixels, respectively. We also computed the mean and variance for $R$, $T$, and $S$ between the two groups on the three levels, as shown in Table 2. We conducted the statistical tests using a linear mixed model (LMM) with the R-style formula, e.g., "$T \sim$ Group + (1 | PromptID) + (1 | UserID)" for $T$, showing again the significant differences between the two groups (with the significance level $p<0.001$) on $R$, $T$, and $S$ over the three levels. It proves again that the novice users could not perform well to organize strokes from global to local or draw a correct path of each stroke to depict a target shape as the professional users did.

### 3.3 Could Scaffold Lines Help Improve Sketches?

Table 4 of the main paper shows that scaffold lines could significantly reduce global errors on scaling $E_{GS}$ and pixel-level errors $E_P$, for both the novice and professional users. On the stroke level, scaffolding made a significant difference on locating stroke positions (see $E_{LT}$) for the novice drawings, but not for the professional drawings. In addition, it shows that the scaffold lines did not successfully guide the orientation of neither the whole sketch nor each stroke for the two groups. We further tested Spearman correlation coefficients $r$ with significant level $p$ between the scaffold-line numbers and the pixel-level error $E_P$ for the two groups using scaffold lines. We found that there was a significantly negative correlation for the novice drawings ($r$=-0.18, $p$<0.001) with scaffold lines, meaning that more scaffold lines resulted in less $E_P$ error, while no significant correlation existed in the professional drawings ($r$=-0.05, $p$=0.11).

### 3.4 Do People Sketch 3D Shape Differently over Time?

We also computed Spearman correlation coefficient between different drawing features and time for each drawing, including stroke length, the speed of drawing each stroke, stroke duration, and the distance between the two end points of each stroke. Figure 4 shows most of the drawings have no significant correlation between the features and time, both for the novice and professional groups.

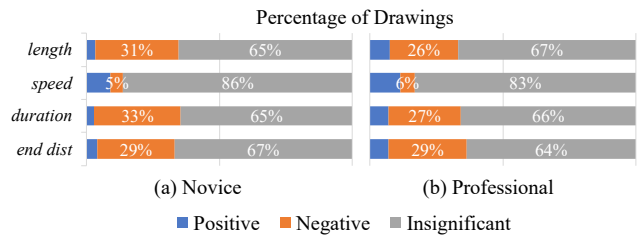

Fig. 4. Histograms of the correlations between different drawing features and time, including stroke length, the speed of drawing each stroke, stroke duration, and the distance between the two end points of each stroke.

We computed the energy costs of the guidelines for each drawing, following Fu et al. [2011]:

$$E_{sim} = \frac{1}{n} \sum_{i}^{n} c_{ind}(l_i)\theta(i), \quad (1)$$

$$E_{pro} = \frac{1}{n-1} \sum_{i}^{n-1} c_{pro}(l_i, l_{i+1}), \quad (2)$$

$$E_{col} = \frac{1}{n-1} \sum_{i}^{n-1} c_{col}(l_i, l_{i+1}), \quad (3)$$

$$E_{anc} = \frac{|\{(l_c, l_s) \in T | c > s\}|}{|T|}, \quad (4)$$

where $l_i$ indicates the $i$-th drawn stroke and $n$ is the stroke count of a drawing, while $T$ is a set of the detected T-junctions $(l_c, l_s)$ between the crossbar stroke $l_c$ and the stem stroke $l_s$. For the anchoring cost $E_{anc}$, we computed the proportion of the T-junctions that were drawn without following the anchoring guideline (i.e., the crossbar $l_c$ is assumed to be drawn earlier than the stem $l_s$, that is, $c<s$). Please refer to [Fu et al. 2011] for the detailed definitions for $c_{ind}(\cdot)$, $\theta(\cdot)$, $c_{pro}(\cdot)$, $c_{col}(\cdot)$, etc.

### 3.5 Stroke-level Comparison

We defined stroke precision as the case that over half of pixels of one stroke are aligned with targets. To observe the difference of the sketching performance on strokes between the novice users and professional users, we computed stroke precision on the stroke-level registered results by setting tracings (we also collected for our dataset) as targets. Figure 5 shows users' performance on each category. Professional users outperformed at all the categories, of which *Human Face*, *Vehicle*, *Animal* have larger disparity between the two groups than *Primitive* and *Lamp* show. It reflects the professional users have better command of sketching complicated 3D objects than novice users do.

## 4 APPLICATION

### 4.1 Freehand-style Sketch Synthesis

We trained the three disturbers using three MLPs (100 epochs), each of which consists of two hidden layers (each one following a ReLU activation layer) respectively with 100 and 50 neurons, using the Adam solver with the fixed learning rate 1e-3. We trained each MLP separately on the novice and professional drawings and the corresponding registered results. The dataset for the novice style consists ∼70k paired strokes while that for the professional style has ∼102k pairs we set a training/testing ratio of the dataset to 20:1 in our experiments.

*Stroke Extrinsic Disturbing*. The input $S_e$ of the extrinsic disturber is a fitted Bézier curve with six control points $t \in \mathbb{R}^{12}$ for each stroke of tracings with the noise level $n_1$, i.e., $S_e \in \mathbb{R}^{13}$, while the output $O_e$ is the transformation parameters for rotation, translation and scaling, i.e., $O_e \in \mathbb{R}^4$. When training, the noise level $n_1$ is computed using the stroke-level errors (mentioned in Section 5.3 of the main paper) by $E_{LR}+E_{LT}+E_{LS}$. Note that each error is normalized before training. When testing, the curve $t$ is applied by the predicted transformation $O_e$ and then is fed into the next intrinsic disturber.

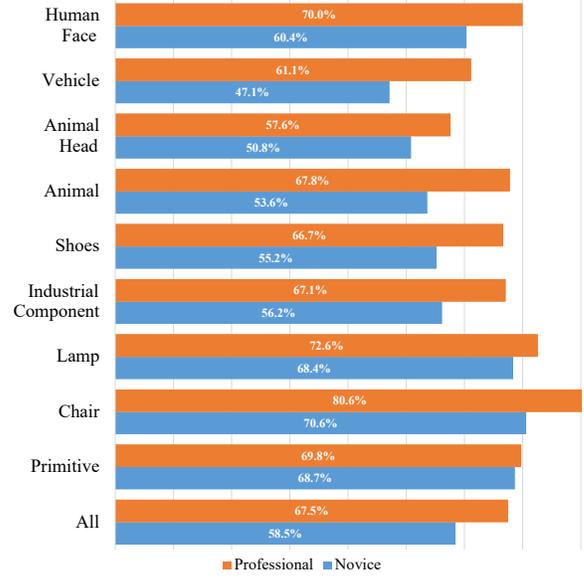

Fig. 5. Stroke precision of the stroke-level registered results for each category.

*Stroke Intrinsic Disturbing*. The input $S_i \in \mathbb{R}^{13}$ and output $O_i \in \mathbb{R}^{12}$ of the intrinsic disturber are both a Bézier curve, while the input is also incorporated with the noise level $n_2$. We compute $n_2$ by measuring the Euclidean distance between $S_i$ and $O_i$ for the training dataset. When testing, the predicted curve $O_i$ is set as an input of the point disturber.

*Point Disturbing*. For the point disturber, the input $S_P \in \mathbb{R}^{12}$ is a Bézier curve and the output $S_P \in \mathbb{R}^2$ is a normal distribution $N(\mu, \sigma)$. When training, we measure the offsets between all the points of the Bézier curve and the original stroke (we use the sketch-level registered results for training) and compute the mean $\mu$ and standard deviation $\sigma$ of the offsets. When testing, the input curve $S_P$ is added with the predicted normal noise before the next step (layout refinement).

We provide more examples for each step (Figure 6) and with different noise levels (Figure 7) to show the plausibility and diversity of our freehand-style synthesis method.

### 4.2 Preliminary Evaluation of Sketch-based 3D Reconstruction

We chose four popular image-based shape reconstruction methods that can take single-view sketches as input. Each of the four chosen methods uses one of the commonly used shape representations: meshes, voxels, implicit functions, and point clouds. First, we chose Pixel2Mesh [Wang et al. 2018] as a representative mesh-based approach, which deforms a template ellipsoid to a target shape. Pixel2Mesh is often adopted as a baseline due to its generality and flexibility on shape reconstruction. The well-known 3D-R2N2 [Choy et al. 2016] network was chosen as a representative volumetric approach for shape reconstruction. We chose Occupancy Network

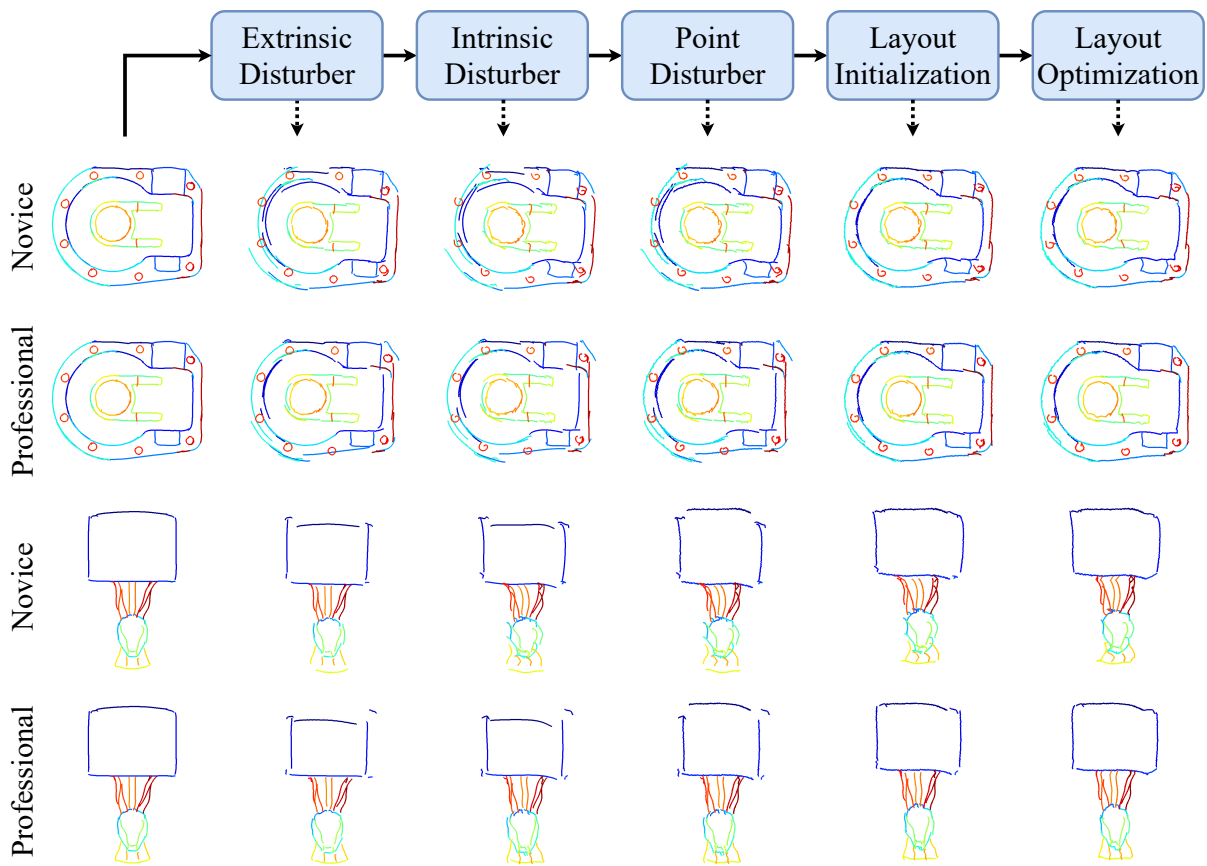

Fig. 6. The pipeline of our freehand-style sketch synthesis method. Each two rows of images show novice-style and professional-style generation, both of which are fed with the same noise levels ($n_1=n_2$). Each stroke of the sketches is color-coded to highlight their changes.

[Mescheder et al. 2019] and PSGN [Fan et al. 2017] as the representative methods based on implicit functions and point clouds, respectively. For a fair comparison, we trained all the four models with synthetic sketches from [Zhong et al. 2020]. Different from the other three methods, which require only a single sketch as input, Pixel2Mesh requires an input sketch and its view angle as input for the subsequent reconstruction. We thus provided the corresponding view angle information paired with the synthetic sketch in training and applied the view angles associated with the reference image as the corresponding view information of freehand sketches in test.

We provide more visual comparisons of the reconstructed shapes given different input freehand sketches in Figure 8. Since Pixel2Mesh reconstructs shapes by deforming an ellipsoid, the resulting models often present excessive adhesion in the leg areas. Due to the limited sampling resolution of voxels, R2N2 presents results with disconnected parts (e.g., legs and back area) in some cases. In addition, due to the adoption of a volumetric representation, R2N2 produces shapes with surfaces expanding the stacks of individual cubic voxels, which also explains its relatively low NC values in the quantitative evaluation. Since the sketch inputs contain only binary shape information, such scarce information would add more difficulties for reconstruction methods to identify the within/outside shape relation, resulting in messy generation for complex parts (e.g., hollow backs and swivel bases).

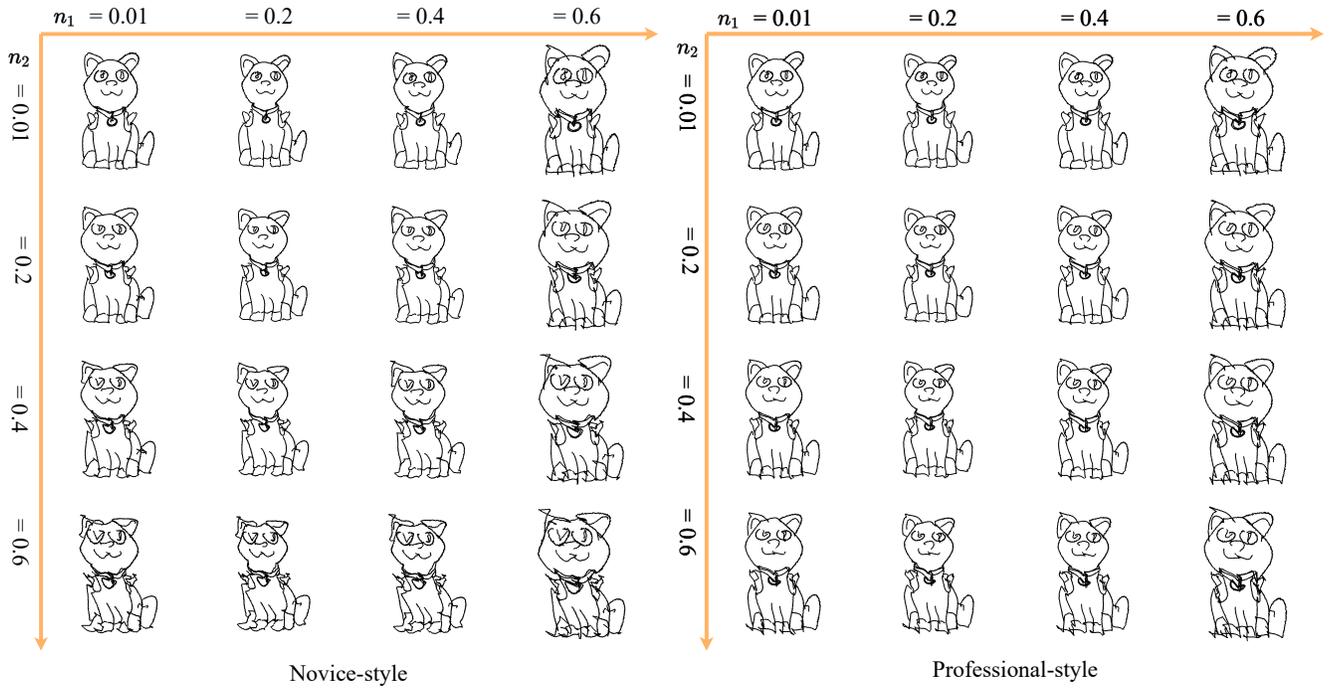

Fig. 7. Examples of our freehand-style synthesis method fed with different noise levels $n_1$ and $n_2$.

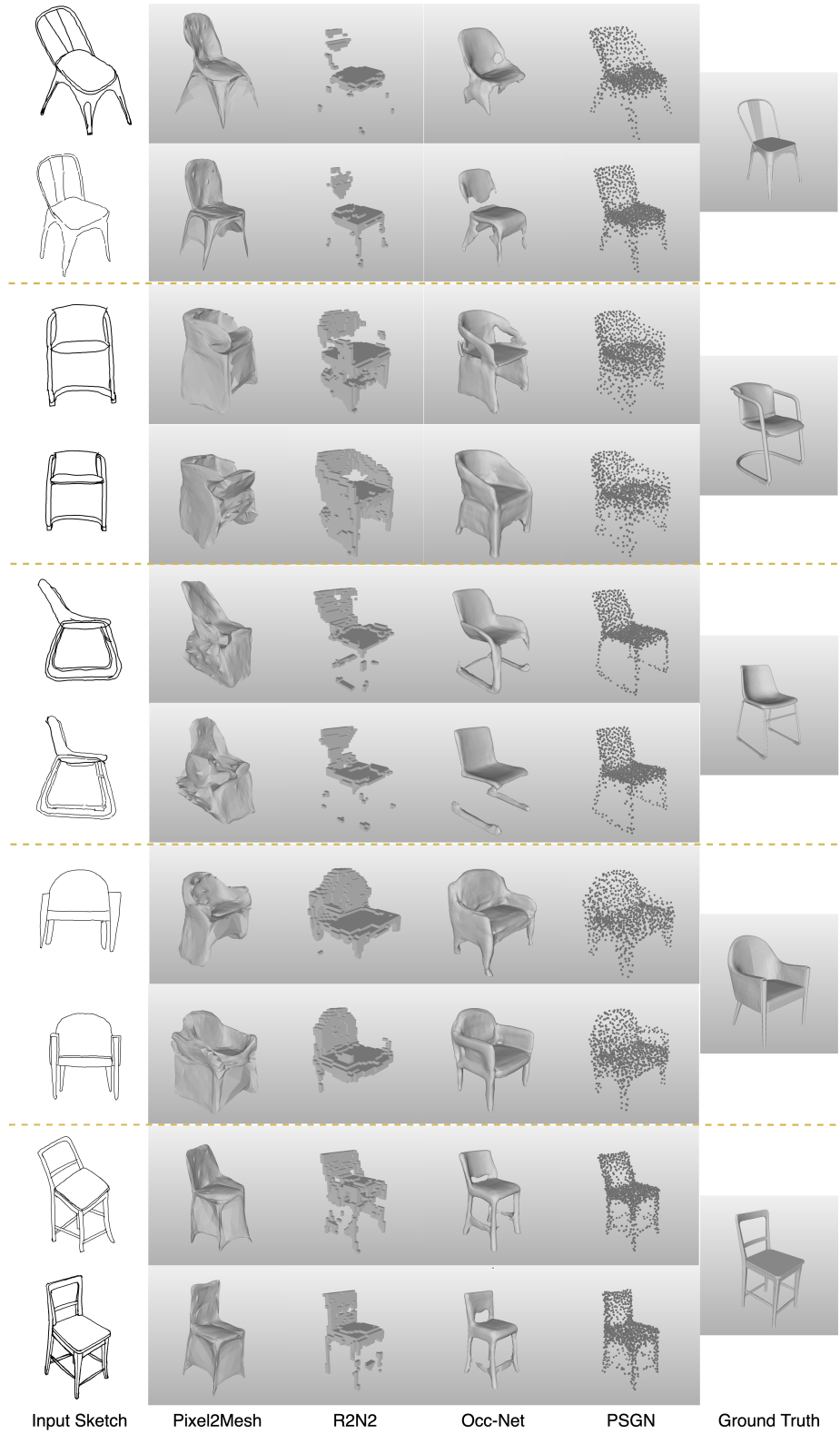

Fig. 8. More results of the 3D reconstruction comparison. In each group, the top presents the results from novice input, the bottom illustrates the corresponding reconstruction results given professional sketches; on the right, we present the ground truth shapes.


\end{document}